\documentclass[10pt,twocolumn,letterpaper]{article}

\usepackage[pagenumbers]{iccv} 

\usepackage[accsupp]{axessibility}

\usepackage{multirow}
\usepackage{colortbl}
\usepackage{makecell}

\definecolor{iccvblue}{rgb}{0.21,0.49,0.74}
\usepackage[pagebackref,breaklinks,colorlinks,allcolors=iccvblue]{hyperref}

\def\paperID{7613} 
\def\confName{ICCV}
\def\confYear{2025}

\title{PropVG: End-to-End Proposal-Driven Visual Grounding with Multi-Granularity Discrimination}

\author{Ming Dai$^{1}$ \ \ Wenxuan Cheng$^1$ \ \ Jiedong Zhuang$^2$ \ \ Jiang-jiang Liu$^3$ \ \ \\ Hongshen Zhao$^1$ \ \ Zhenhua Feng$^4$ \ \ Wankou Yang$^{1*}$ \\
$^1$Southeast University \ \ $^2$ Zhejiang University \ \ $^3$Baidu VIS \ \  $^4$ Jiangnan University
}

\definecolor{ggray}{RGB}{127,127,127}
\definecolor{aliceblue}{rgb}{0.94, 0.97, 1.0}

\begin{document}
\maketitle

\begin{abstract}
    Recent advances in visual grounding have largely shifted away from traditional proposal-based two-stage frameworks due to their inefficiency and high computational complexity, favoring end-to-end direct reference paradigms. However, these methods rely exclusively on the referred target for supervision, overlooking the potential benefits of prominent prospective targets. Moreover, existing approaches often fail to incorporate multi-granularity discrimination, which is crucial for robust object identification in complex scenarios. To address these limitations, we propose \textbf{PropVG}, an end-to-end proposal-based framework that, to the best of our knowledge, is the first to seamlessly integrate foreground object proposal generation with referential object comprehension without requiring additional detectors. Furthermore, we introduce a \textit{Contrastive-based Refer Scoring} (CRS) module, which employs contrastive learning at both sentence and word levels to enhance the model’s capability in understanding and distinguishing referred objects. Additionally, we design a \textit{Multi-granularity Target Discrimination} (MTD) module that fuses object- and semantic-level information to improve the recognition of absent targets. 
    Extensive experiments on gRefCOCO (GREC/GRES), Ref-ZOM, R-RefCOCO/+/g, and RefCOCO/+/g (REC/RES) benchmarks demonstrate the effectiveness of PropVG. The codes and models are available at \url{https://github.com/Dmmm1997/PropVG}.
\end{abstract}    
\section{Introduction}
\label{sec:intro}

\begin{figure}
    \centering
    \includegraphics[width=1.0\linewidth]{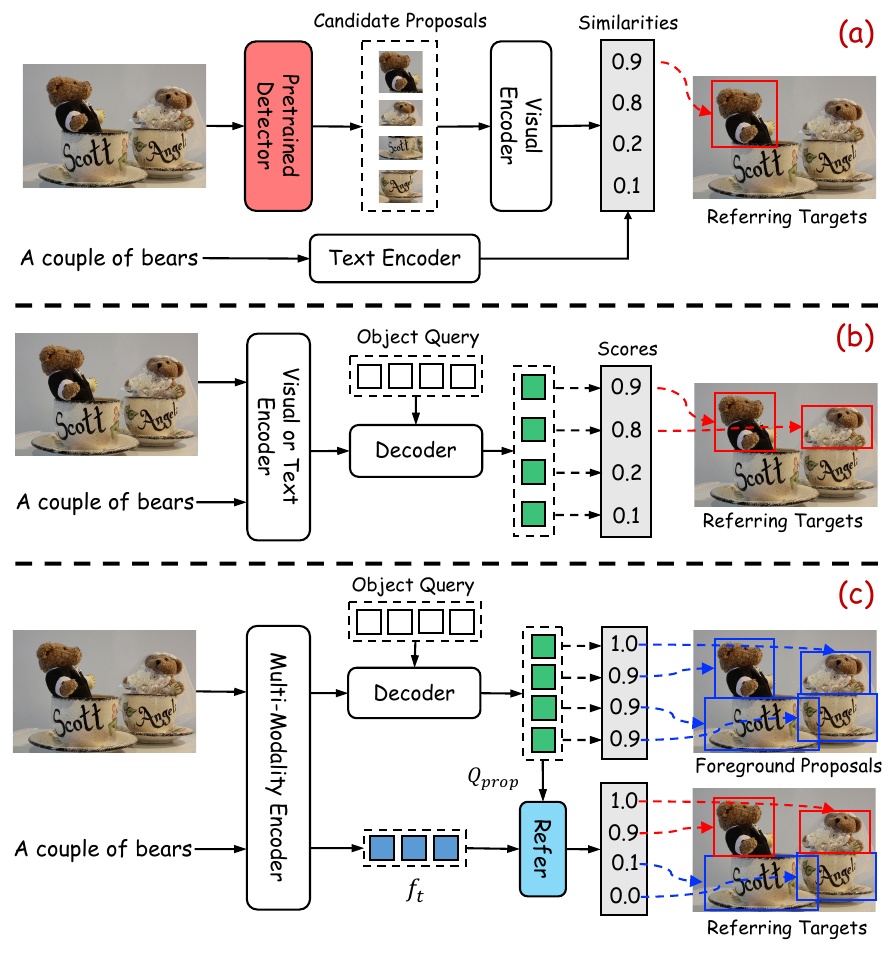}
    \vspace{-20pt}
    \caption{\textbf{Architecture Comparison.} (a) The traditional proposal-based framework relies on pre-trained detectors for proposal generation. (b) The query-based direct referring framework. (c) The proposed \textbf{PropVG} framework, which adopts an end-to-end, detector-free design. It first generates region candidates in the proposal stage and subsequently establishes associations with the referring expression.}
    \label{fig:motivation}
   \vspace{-10pt}
\end{figure}

Visual Grounding (VG)~\cite{glip} aims to localize and segment objects in images based on natural language queries. 
Classic VG tasks mainly encompass Referring Expression Comprehension (REC)~\cite{hu2016natural,resc,transvg,scanformer} and Referring Expression Segmentation (RES)~\cite{husegmentation2016,vlt,magnet}. They primarily focus on scenarios where a single sentence corresponds to a single target. Recently, Generalized Visual Grounding (GVG), which includes Generalized Referring Expression Comprehension/Segmentation (GREC~\cite{grec}/GRES~\cite{rela}), extends classic VG by accommodating scenarios involving zero or multiple referred objects. 
Unlike classic VG tasks, which only require the localization of the most relevant position, GVG demands that the model comprehensively understand the contextual relationship between the image and the expression, accurately identifying all positions corresponding to the referring expression, while also accounting for the possibility of non-existent targets.

Traditional two-stage visual grounding approaches~\cite{cmn, mattnet, cmatterase, nmtree}, as illustrated in Fig.~\ref{fig:motivation}(a), rely on pre-trained detectors to generate candidate proposals~\cite{fasterRCNN,maskRCNN}, which are then matched with expressions. However, this paradigm has fallen out of favor due to its suboptimal performance and high computational complexity. In response, recent methods~\cite{qrnet, seqtr, lads, simvg} adopt direct referring frameworks that optimize solely for the referred target's location, simplifying the training process of visual grounding. While the optimization objective of this framework is clear, it inevitably weakens the model’s general perceptual localization ability. By focusing exclusively on the referred target, the model's interpretability of the referentiality of other high-response targets diminishes significantly. On the other hand, object existence discrimination in generalized scenarios presents another significant challenge. Existing methods~\cite{grec, rela, mabp}, as shown in Fig.~\ref{fig:motivation}(b), address this problem from the perspective of global or single-granularity predictions. However, these approaches overlook the importance of multi-granularity predictions in accurately determining object existence.

To address these challenges, we propose an end-to-end proposal-based framework, PropVG, which is specifically designed to overcome the limitations of previous proposal-based approaches~\cite{mattnet,ref-nms}, particularly their suboptimal performance and slow inference speed. Unlike direct referring methods~\cite{yang2019fast, seqtr, rela, simvg}, PropVG not only effectively localizes referred targets but also concurrently enables the perception of prominent foreground objects. Furthermore, it is extendable to generalized visual grounding tasks~\cite{grec, rela, rris, refzom}, enhancing both its versatility and practical applicability. In summary, PropVG addresses two key challenges: \textit{\textbf{(1)} how to leverage a proposal-based framework to ensure precise identification of referred targets while facilitating effective foreground object perception without relying on pre-trained detectors; \textbf{(2)} how to improve object existence discrimination in generalized scenarios.}

\textbf{\textit{Contrastive-based End-to-end Proposal-based Framework.}} We propose an end-to-end proposal-based visual grounding architecture that integrates a DETR-type detector. The framework consists of two primary stages: foreground proposal generation and target referentiality assessment, as illustrated in Fig.~\ref{fig:motivation}(c). In the first stage, a multi-scale deformable decoder~\cite{deformabledetr} is employed to generate foreground proposals. The second stage introduces a Contrastive-based Refer Scoring (CRS) module, which evaluates the referentiality of each proposal. Specifically, the CRS module is designed to assess the relevance between the proposals and the referring expressions, enabling the model to accurately identify the most relevant target. By combining foreground perception with referents localization in a unified end-to-end design, PropVG overcomes the underperformance and inefficiencies traditionally associated with previous proposal-based methods~\cite{mattnet,nmtree,ref-nms}, revitalizing the potential of this architecture.

\textbf{\textit{Multi-granularity Object Existence Discrimination.}} We propose a multi-task collaborative framework for generalized scenarios, incorporating a Multi-granularity Target Discrimination (MTD) module. This module introduces a score prior cross attention mechanism that integrates prior score distribution information into the attention map. Additionally, we directly inject the predicted values of referring and segmentation scores to ensure consistency in target existence across multiple predictions. This approach enables MTD to effectively combine semantic- and object-level granularities, thereby enhancing global target existence judgment.

The main contributions are summarized as follows:
\begin{itemize}
    \item We propose an end-to-end multi-task collaborative framework, named \textit{PropVG}, which is the first proposal-based framework without relying on pre-trained detectors.
    \item We introduce a contrastive-based refer scoring module, which adaptively balances and integrates the contributions of sentence- and word-level contrastive learning to effectively assess the referentiality of proposals. Additionally, we integrate a multi-granularity target discrimination module, which enhances target existence recognition by incorporating semantic and object-level priors.
    \item The proposed PropVG overcomes the performance and inference latency limitations of traditional proposal-based methods (4$\times$ faster, +14\% Acc.), achieving significant improvements over advanced approaches on 10 datasets.
\end{itemize}

\section{Related Work}
\label{sec:releatedwork}

\subsection{Referring Expression Comprehension (REC) / Segmentation (RES)} 
In REC, each sentence corresponds to a unique target bounding box in the image. Early studies primarily focused on two-stage approaches~\cite{cmn,vc,parallelattention,mattnet,cmatterase,dga,ddpn,rvg-tree,cm-att-erase,ref-nms}, where candidate proposals are first generated using pre-trained detectors, followed by matching the referring expressions to these proposals to identify the most relevant target. Later, one-stage methods~\cite{realgin,mcn,resc} adopted dense anchor strategies~\cite{yolov3}, improving both inference efficiency and performance. 
Recently, numerous Transformer-based methods~\cite{transvg, transvg++, mdetr, dynamicmdetr, dqdetr, scanformer, simvg, oneref} have been proposed to effectively capture cross-modal relationships, significantly enhancing multimodal understanding. Meanwhile, some research~\cite{dara,mapper,detris} has explored parameter-efficient tuning strategies to reduce the computational cost of training.

In RES, each sentence is associated with a set of pixels that delineate the target object in the image. Traditional approaches~\cite{husegmentation2016, liu2017, cmpc, efn} predominantly rely on convolution-based operations for cross-modal fusion to generate segmentation masks. However, these methods struggle to effectively capture complex visual-linguistic relationships. To address this limitation, recent studies~\cite{lavt, vlt, mmm, cris, restr, lqmformer, magnet, mmca, falip, DeRIS} have incorporated attention-based mechanisms~\cite{transformer} to enhance multi-modal interactions.  
Another line of research~\cite{mcn, reftr, seqtr, polyformer, pvd, vg-law, eevg, c3vg} focuses on multi-task joint learning, aiming to unify REC and RES to reduce redundant computation and parameter overhead. With the advent of Multimodal Large Language Models (MLLMs)~\cite{llama, llava, zhuang2025st3, li2024survey}, several methods~\cite{kosmos, groundinggpt, lion, lisa, gsva, glamm, sam4mllm} have started leveraging the powerful image-text understanding capabilities of MLLMs to achieve more generalizable visual perception.

In this paper, we further investigate the effectiveness of two-stage proposal-based architectures. Diverging from traditional approaches that depend on pre-trained detectors, we partition the model into two discrete stages (proposal generation and target discrimination) for end-to-end training.

\subsection{Generalized Visual Grounding (GVG)}

GVG extends the classic VG task by addressing scenarios involving empty or multiple targets. It can be further divided into two subtasks: Generalized Referring Expression Comprehension (GREC)~\cite{grec} / Segmentation (GRES)~\cite{rela}. DMMI~\cite{refzom} introduced a new benchmark for beyond-single-target segmentation, while RefSegformer~\cite{rris} enhanced transformer-based models with non-target sentence discrimination, leading to more robust segmentation performance.  
More recently, ReLA~\cite{rela} proposed the Generalized RES task, expanding the scope to accommodate both empty and multiple-target cases. Building upon this, GREC~\cite{grec} further extended GRES from segmentation to detection tasks. Although these methods leverage Transformer-based~\cite{transformer} architectures to directly predict referred targets, they overlook the potential benefits of utilizing foreground objects to enhance perception capabilities.

In this paper, we develop a multi-task collaborative framework and design a multi-granularity target discrimination module, which incorporates both object- and semantic-level information to determine the presence of targets.

\begin{figure*}
    \centering
    \includegraphics[width=1.0\linewidth]{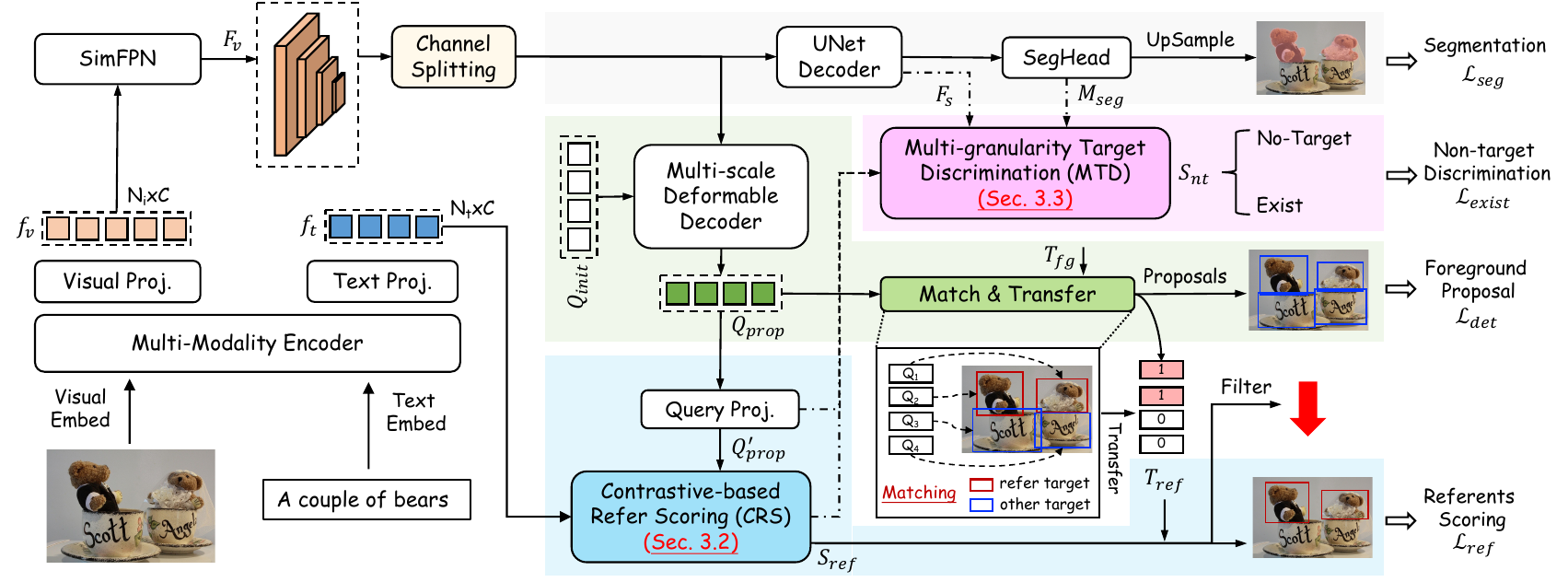}
    \vspace{-15pt}
    \caption{\textbf{Overview of PropVG.} PropVG utilizes the BEiT-3~\cite{beit3} architecture to extract multi-modal representations, followed by the SimFPN module for multi-scale feature extraction. The framework then branches into two pathways: one employs a UNet Decoder and SegHead for global referring segmentation, while the other generates foreground proposals using a Multi-scale Deformable Decoder and DetHead. The CRS module computes the relevance score for each query, and the MTD module evaluates target existence.}
    \label{fig:framework}
    \vspace{-5pt}
\end{figure*}

\section{The Proposed PropVG Framework}
\label{sec:method}

In this section, we begin by presenting an overview of the PropVG framework in Sec.~\ref{subsec:overview}. Subsequently, we introduce the contrastive-based refer scoring module in Sec.~\ref{subsec:crs}, followed by the multi-granularity target discrimination module in Sec.~\ref{subsec:mtd}. Last, we describe the training objectives in Sec.~\ref{subsec:objective_target}. Additional introduction regarding dataset construction and post-processing is provided in Appendix~C.

\subsection{Overview}
\label{subsec:overview}
As illustrated in Fig.~\ref{fig:framework}, the process begins by independently embedding and processing an image $\mathcal{I} \in \mathbb{R}^{H_v \times W_v \times 3}$ and a textual expression $\mathcal{T}$ using a multi-modality encoder (BEiT-3~\cite{beit3}). This encoder performs vision-language encoding and fusion, producing both visual and textual features.
Next, we apply image projection and text projection to map them to a lower-dimensional space $C$, resulting in $f_v \in \mathbb{R}^{N_i \times C}$ and $f_t \in \mathbb{R}^{N_t \times C}$. 
Subsequently, we extend the single-scale visual features $f_v$ into multi-scale feature maps, denoted as $f_v^j \in \mathbb{R}^{H_j \times W_j \times C_j}$, where $j \in {0, 1, 2, 3}$, leveraging the SimFPN architecture~\cite{vitdet}. These multi-scale features are then divided along the channel dimension into two distinct subsets, each tailored to support the segmentation and detection tasks, respectively.

\textit{For segmentation}, a UNet decoder~\cite{unet} integrates the multi-scale features, and a SegHead predicts the global segmentation mask $M_{seg}$. The result is then upsampled to the original resolution for pixel-level supervision.

\textit{For detection}, we decompose the process into two distinct components: foreground proposal and referent scoring. (1) In the foreground proposal phrase, we initialize $N$ learnable queries $Q_{init}$. These queries interact with multi-scale features through a multi-scale deformable decoder, resulting in refined proposal queries $Q_{prop} \in \mathbb{R}^{N \times C}$. Subsequently, a lightweight detection head (DetHead) generates foreground bounding box proposals $P_{bbox} \in \mathbb{R}^{N \times 4}$ along with their corresponding confidence scores $P_{score} \in \mathbb{R}^{N \times 2}$. To establish correspondence between queries and targets, we apply the Hungarian matching algorithm~\cite{detr} to align the queries with the all ground-truth foreground objects, yielding the query-to-target assignment set $T_{ref}$ for use in the referring phrase.
(2) In the referent scoring phase, a Query Proj. module maps $Q_{prop}$ to $Q_{prop}'$, thereby adapting the queries for the query-level referring classification task. A contrastive-based referent scoring module then calculates referring scores for each query, with supervision provided by the assignment set $T_{ref}$. Furthermore, to accommodate generalized scenarios necessitating the determination of target presence, we propose a multi-granularity target discrimination module. This module integrates both object-level and semantic-level information to comprehensively evaluate the existence of the target, enhancing the model’s adaptability and robustness across diverse contexts.

\subsection{Constrastive-based Refer Scoring}
\label{subsec:crs}

\begin{figure}
    \centering
    \includegraphics[width=1.0\linewidth]{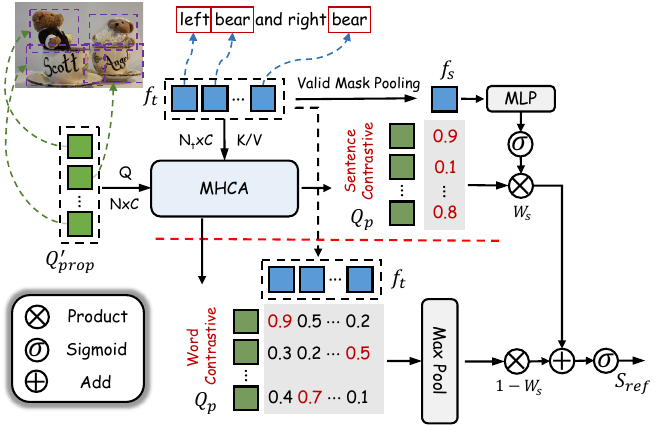}
    \vspace{-15pt}
    \caption{\textbf{Architecture of CRS.} The architecture begins with a MHCA to facilitate interaction between $Q_{prop}'$ and $f_v$. Subsequently, sentence- and word-level contrastive learning are applied to compute the referring score. Finally, a learnable weight parameter is introduced to balance the contributions of these two levels.}
    \label{fig:CRS}
    \vspace{-15pt}
\end{figure}

This module is designed to compute the relevance score for each query. As shown in Fig.~\ref{fig:CRS}, it follows a contrastive learning paradigm, combining sentence- and word-level query-to-text similarities through adaptive weighting. 
First, the Multi-Head Cross-Attention (MHCA) mechanism enables interaction between queries and textual features, generating text-aware query features $Q_p$. Subsequently, sentence-level contrastive learning computes the similarity matrix $S_{sent} \in \mathbb{R}^{N \times 1}$ between $Q_p$ and the global textual features $f_s \in \mathbb{R}^{1 \times C}$. The global features $f_s$ are obtained by applying a valid mask pooling operation over the word-level textual features $f_t$ as follows:
\begin{equation}
    \setlength{\abovedisplayskip}{5pt}
    \setlength{\belowdisplayskip}{5pt}
    \begin{aligned}
        f_{s}^{i} = \max \left[ f_{t}^{i} \times (\sim m) \right], \quad i \in \{1,2,...,C\},
    \end{aligned}
\end{equation}
where $m$ is the padding mask of text tokens. This operation preserves the maximum response of each token across channels by selecting the highest value along the token dimension. The similarity matrix is computed as:
\begin{equation}
    \setlength{\abovedisplayskip}{5pt}
    \setlength{\belowdisplayskip}{5pt}
    \begin{aligned}
        \text{Sim}(f_1, f_2) = \frac{f_1 \cdot f_2}{\|f_1\| \|f_2\|} / T,
    \end{aligned}
\end{equation}
where $T$ is a learnable scaling factor initialized to 0.07.  
Meanwhile, word-level contrastive learning computes the similarity between query features and word-level textual features $f_t$, producing a similarity matrix $S_{word} \in \mathbb{R}^{N \times N_t}$. A learnable weight is used to dynamically balance the contributions of the sentence- and word-level branches based on the global textual features. Specifically, the weight $w_s$ is obtained by applying a MLP followed by a sigmoid activation to $f_s$, which determines the allocation to the sentence-level branch. The final referring score for each query is computed as:
\begin{equation}
    \setlength{\abovedisplayskip}{5pt}
    \setlength{\belowdisplayskip}{5pt}
    \begin{aligned}
        S_{ref} = w_s \cdot S_{sent} + (1 - w_s) \cdot\text{MaxPool}(S_{word}).
    \end{aligned}
\end{equation}
This design adaptively balances and integrates the contributions of sentence- and word-level contrastive learning.

\subsection{Multi-granularity Target Discrimination}
\label{subsec:mtd}

\begin{figure}
    \centering
    \includegraphics[width=1.0\linewidth]{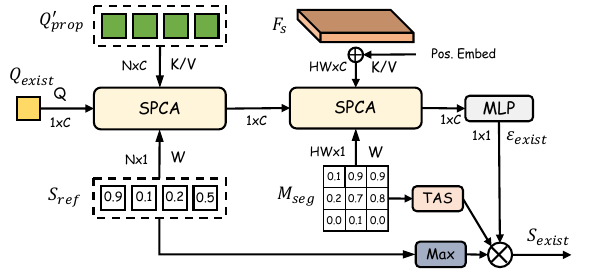}
    \vspace{-10pt}
    \caption{\textbf{Architecture of MTD.} The MTD comprises two SPCA blocks to integrate object- and semantic-level predictions, using the referring score $S_{ref}$ and segmentation prediction $M_{seg}$ as attention priors. TAS calculates the average score of the top $K$ pixels in the segmentation prediction.}
    \label{fig:MTD}
   \vspace{-10pt}
\end{figure}

The MTD module first predefines a learnable query $Q_{exist}$, which then integrates both object- and semantic-level features through two custom-designed Score Prior Cross Attention (SPCA) modules. The primary purpose of SPCA is to leverage both the refer score and semantic score to provide query- and pixel-level prior information, facilitating accurate target existence estimation. Compared to the standard attention mechanism, SPCA is formulated as:
\begin{equation}
    \label{eq:spca}
    \setlength{\abovedisplayskip}{5pt}
    \setlength{\belowdisplayskip}{5pt}
    \begin{aligned}
        O = \text{Softmax}(QK^{T} + \text{MLP}(S))V,
    \end{aligned}
\end{equation}
where $S$ represents the weights (denoted as $W$ in Fig.~\ref{fig:MTD}), including $S_{ref}$ and $M_{seg}$. A MLP is employed to adjust $S$ to align with the distribution range of the attention. After interacting with both object- and semantic-level priors, the $Q_{exist}$ is mapped to a scalar, representing the probability of target existence, denoted as $\varepsilon_{exist}$.
To further enhance the accuracy of target existence determination, we explicitly incorporate both the $S_{ref}$ and $M_{seg}$. The final target existence score $S_{exist}$ is computed as:
\begin{equation}
    \setlength{\abovedisplayskip}{5pt}
    \setlength{\belowdisplayskip}{5pt}
    \begin{aligned}
        S_{exist} = \text{Max}(S_{ref}) \times \text{TAS}(M_{seg}) \times \varepsilon_{exist},
    \end{aligned}
\end{equation}
where $\text{TAS}$ denotes the TopK Average Score, which calculates the average score of the top $K$ pixels in the segmentation prediction. This design mitigates the influence of high-confidence outliers in segmentation predictions while preserving the global information from the segmentation mask.

\subsection{Training Objectives}
\label{subsec:objective_target}

The training objectives consist of four components: detection, referring, global segmentation, and target existence.
\textbf{Detection}: The detection branch employs a loss function $\mathcal{L}_{detr}$ similar to that of DETR~\cite{detr}, combining L1, Cross-Entropy, and GIoU loss functions to address the detection task.
\textbf{Referring Branch}: The detection branch generates proposals for foreground objects, and the referring branch determines whether these proposals correspond to the target referred to by the given text. A Binary Cross-Entropy (BCE) loss is used to measure the deviation between the referring score $S_{ref}$ and the ground truth $T_{ref}$.
\textbf{Global Segmentation}: This branch uses the BCE and Dice loss functions, following the approach in~\cite{cris}, to quantify the difference between the predicted global mask $S_{global}$ and the gt mask $\mathcal{M}_{gt}$.
\textbf{Target Existence Branch}: This branch handles binary classification by determining the presence of the target across the global scene. It utilizes the BCE loss to classify whether the target exists.
Finally, the total loss can be defined as:
\begin{equation}
    \setlength{\abovedisplayskip}{5pt}
    \setlength{\belowdisplayskip}{5pt}
    \begin{aligned}
    \mathcal{L}_{total} &=  \mathcal{L}_{seg} + \lambda_{det} \cdot \mathcal{L}_{det} \\
    &\quad + \lambda_{exist} \cdot \mathcal{L}_{exist} + \lambda_{ref} \cdot \mathcal{L}_{ref},
    \end{aligned}
    \label{eq:loss}
\end{equation}
by default, $\lambda_{det} = 0.1$, $\lambda_{exist} = 0.2$, and $\lambda_{ref} = 1.0$.

\section{Experimental Results}
\label{sec:experiment}
In this section, we first provide a brief overview of the experimental setup (Sec.~\ref{subsec:experiment_setup}). Then, we compare our method with SOTA approaches (Sec.~\ref{subsec:main_results}). Last, we conduct ablation studies on the proposed method (Sec.~\ref{subsec:ablation}) and present some qualitative results (Sec.~\ref{subsec:visualization}). Additional experiments are provided in Appendix~E.

\subsection{Experimental Setups}
\label{subsec:experiment_setup}
We evaluate our model on ten datasets: gRefCOCO (GREC~\cite{grec}/GRES~\cite{rela}), R-RefCOCO/+/g~\cite{rris}, Ref-ZOM~\cite{refzom} and RefCOCO/+/g~\cite{refcoco/+,refcocog} (REC/RIS). Limited to the space, the descriptions of datasets and metrics are elaborated in Appendix A and B. The implementation details are described in Appendix D.

\subsection{Main Results}
\label{subsec:main_results}

\begin{table*}[t]
    \setlength{\tabcolsep}{3pt}
    \renewcommand{\arraystretch}{1.18}
    \footnotesize
    \begin{center}
        \resizebox{0.94\linewidth}{!}{
        \begin{tabular}{l|c|c|ccc|ccc|ccc|ccc|cc|cc|c}
            \hline
            \multirow{3}{*}{Models} & \multirow{3}{*}{\makecell{Visual \\ Encoder}} & \multirow{3}{*}{Pretrain} & \multicolumn{6}{c|}{RefCOCO} & \multicolumn{6}{c|}{RefCOCO+} & \multicolumn{4}{c|}{RefCOCOg} & \multirow{3}{*}{\makecell{Time \\ (ms)}} \\
            & & & \multicolumn{3}{c|}{REC} & \multicolumn{3}{c|}{RES} & \multicolumn{3}{c|}{REC} & \multicolumn{3}{c|}{RES} & \multicolumn{2}{c|}{REC} & \multicolumn{2}{c|}{RES} & \\
            & & & val & testA & testB & val & testA & testB & val & testA & testB & val & testA & testB & val-u & test-u & val-u & test-u & \\
            \hline
            \rowcolor{ggray!20}
            \multicolumn{20}{c}{\textbf{\textit{MLLM Methods}}} \\
            \hline       
            LISA-L2-13B~\cite{lisa} & SAM-ViT-H & \checkmark &  85.91 & 88.84 & 81.73 & 76.30&78.70&72.40& 74.46 & 80.56 & 68.26 & 66.20&71.00&59.30&  80.09 & 81.27 & 70.10& 71.10&  -  \\
            GSVA-L2-13B ~\cite{gsva} & SAM-ViT-H & \checkmark&  89.16 & 92.08 & 87.17 &79.20&81.70&77.10&79.74 &84.45 &73.41 &70.30&73.80& 63.60& 85.47 &86.18& 75.70 & 77.00 &  - \\
            GLaMM-V-7B ~\cite{glamm} & CLIP-ViT-H & \checkmark& - & - & - &{79.50}&\underline{83.20}&76.90&- &- &- &{72.60}&\underline{78.70}& 64.60& -&-& \underline{74.20} & {74.90} &  - \\
            \hline       
            \rowcolor{ggray!20}
            \multicolumn{20}{c}{\textbf{\textit{Specialist Methods}}} \\
            \hline
            \multicolumn{20}{l}{\textbf{Direct Referring Methods:}} \\
            TransVG~\cite{transvg}& ResNet-101&$\times$&81.02 & 82.72 & 78.35  &-&-&- & 64.82 & 70.70 & 56.94 &-&-&- & 68.67 & 67.73&-&-&62\\  
            LAVT~\cite{lavt} & Swin-B &$\times$& - & - & - &74.46&76.89&70.94& - & - &-&65.81&70.97&59.23&-&-&63.34&63.62& - \\
            VG-LAW~\cite{vg-law} & ViT-B &$\times$&86.62 &89.32 &83.16 &75.62 &77.51& 72.89&76.37 &81.04 &67.50&66.63&70.38 &58.89&76.90 &76.96&65.63 & 66.08& - \\
            SegVG~\cite{segvg}& ResNet-101&$\times$&86.84 & 89.46 & 83.07  &-&-&- & 77.18 & 82.63 & 67.59 &-&-&- & 78.35 & 77.42&-&-&-\\
            DETRIS-B~\cite{detris}& DINOv2-B &$\times$&-&-&-& 76.00 & 78.20 &73.50 &-&-&-&68.90 &74.00 &61.50 &-&-& 67.90 &68.10 &-\\
            \hline        
            SeqTR~\cite{seqtr} & DarkNet-53 &\checkmark& 87.00 & 90.15 & 83.59 & 71.70& 73.31& 69.82& 78.69 &84.51 &71.87 & 63.04& 66.73& 58.97& 82.69& 83.37 & 64.69& 65.74 &  \textbf{50} \\
            GroundingDINO~\cite{groundingdino} & Swin-T &\checkmark& 89.19 &  91.86 & 85.99 &-&-& & 81.09 &87.40 &74.71 &-&-&-& 84.15& 84.94&-&-& 120 \\
            PolyFormer~\cite{polyformer} & Swin-B &\checkmark& 89.73 &  91.73 & 86.03&75.96&77.09& 73.22& 83.73 &88.60 &76.38 &70.65&74.51&64.64& 84.46& 84.96& 69.36&69.88&  150 \\
            EEVG~\cite{eevg}&ViT-B&\checkmark&90.47 & 92.73 &87.72 &79.49&80.87&\underline{77.39}&81.79 &87.80 &74.94 &71.86&76.67&66.31&85.19 &84.72&73.56&73.47&117 \\ 
            HiVG~\cite{hivg}&CLIP-ViT-B &\checkmark& 90.56 &92.55 & 87.23 &-&-&-&83.08 &87.83&76.68&-& -& -& 84.71& 84.69&-&-&  - \\
            SimVG-DB~\cite{simvg} & BEiT3-ViT-B &\checkmark& 91.47 &  93.65 & 87.94 &-&-&-& 84.83 &88.85 &79.12 &-&-&-& {86.30}& {87.26} &-&- & \underline{52} \\
            OneRef~\cite{oneref} & BEiT3-ViT-B &\checkmark& \underline{91.89} &  \underline{94.31} & \underline{88.58} &\underline{79.83}&81.86&76.99& \underline{86.38} &\underline{90.38} &\underline{79.47} &\underline{74.68}&77.90&\underline{69.58}& \underline{86.82}& \underline{87.32} &74.06&\underline{74.92} & - \\
            \hline
            \multicolumn{20}{l}{\textbf{Proposal-based Methods:}} \\
            MAttNet*~\cite{mattnet} & ResNet-101&$\times$& 76.65 & 81.14 & 69.99 &56.51&62.37& 51.70& 65.33  & 71.62 & 56.02 &46.67& 52.39&40.08& 66.58 & 67.27 & 47.64&48.61& 320 \\
            NMTree*~\cite{nmtree} & ResNet-101&$\times$& 76.41 & 81.21 & 70.09 &56.59&63.02&52.06& 66.46 & 72.02 & 57.52 &47.40&53.01&41.56& 65.87 & 66.44 &46.59&47.88& - \\
            Ref-NMS*~\cite{ref-nms} & ResNet-101&$\times$& 80.70 & 84.00 & 76.04  &61.46&65.55 &57.41& 68.25 & 73.68 & 59.42 &49.76 &53.84&42.66& 70.55 & 70.62 &51.21&51.90 & - \\
            \hline
            \rowcolor{aliceblue!90} 
            \textbf{PropVG} & BEiT3-ViT-B  &$\times$& {88.96} &{91.55}& {85.73} & {77.99}& {79.81}& {75.28}& {83.72}& {88.00}& {76.60}& {72.94} & {76.49} & {67.22} &{83.50} & {84.44}&{71.34} &{72.10} & 76 \\
            \rowcolor{aliceblue!40} 
            \textbf{PropVG*} & BEiT3-ViT-B  &$\times$& {-} &{-}& {-} & {76.80}& {79.57}& {73.67}& {-}& {-}& {-}& {70.24} & {74.32} & {63.42} &{-} & {-}&{69.30} &{70.53} & 76 \\
            \rowcolor{aliceblue!90} 
            \textbf{PropVG} & BEiT3-ViT-B  &\checkmark& \textbf{92.70} &\textbf{95.07}& \textbf{89.58} & \textbf{81.96}& \textbf{83.58}& \textbf{80.02}& \textbf{87.27}& \textbf{90.87}& \textbf{81.26}& \textbf{77.14} & \textbf{79.83} & \textbf{72.18} &\textbf{88.15} & \textbf{88.30}&\textbf{76.97} &\textbf{77.72} & 76 \\
            \rowcolor{aliceblue!40} 
            \textbf{PropVG*} & BEiT3-ViT-B &\checkmark& - &-& - & 81.80& 83.74& 79.33& -& -& -& 74.81& 78.72& 69.15& - & - &75.54 &77.40 & 76 \\
            \hline
        \end{tabular}
        }
        \vspace{-5pt}
        \caption{Comparison with the SOTA methods on RefCOCO/+/g (REC/RES)~\cite{refcoco/+,refcocog-umd}. \textbf{-L2-13B} refers to LLaMA2-13B~\cite{llama2}, \textbf{-V-7B} refers to Vicuna-7B. The * indicates that the RES metric corresponds to oIoU, while the others are reported using mIoU.}
        \label{table:sotaonpretrain}
        \vspace{-13pt}
    \end{center}
\end{table*}

\begin{table*}
    \setlength{\tabcolsep}{4pt}
    \centering
    \footnotesize
    \renewcommand\arraystretch{1.15}
        \resizebox{0.7\linewidth}{!}{
        \begin{tabular}{l|c|ccc|ccc|ccc}
        \specialrule{.1em}{.05em}{.05em}
        \multirow{2}{*}{Method} & \multirow{2}{*}{Backbone} & \multicolumn{3}{c|}{Val} & \multicolumn{3}{c|}{TestA}  & \multicolumn{3}{c}{TestB} \\
        & & gIoU & cIoU & N-acc. & gIoU & cIoU & N-acc. & gIoU & cIoU & N-acc.\\
        \hline
        \rowcolor{ggray!20}
        \multicolumn{11}{c}{\textbf{\textit{MLLM Methods}}} \\
        \hline
        LISA-V-7B~\cite{lisa} & SAM-ViT-H & 61.63 & 61.76 & 54.67 & 66.27 & 68.50 & 50.01 & 58.84 & 60.63 & 51.91 \\
        GSVA-V-7B~\cite{gsva} & SAM-ViT-H & 66.47 & 63.29 & 62.43 & 71.08 & 69.93 & 65.31 & 62.23 & 60.47 & 60.56 \\
        SAM4MLLM-QW-7B~\cite{sam4mllm} & SAM-EfficientViT-XL1 & \underline{68.96} & \underline{66.33} & 62.96 &70.54 &70.13 & \underline{65.82} & \underline{63.98} & \underline{63.21} & \underline{61.61} \\
        \hline
        \rowcolor{ggray!20}
        \multicolumn{11}{c}{\textbf{\textit{Specialist Methods}}} \\
        \hline
        \multicolumn{11}{l}{\textbf{Direct Referring Methods:}} \\
        VLT~\cite{vlt} & DarkNet-53 & 52.00 & 52.51 & 47.17 & 63.20 & 62.19 & 48.74 & 50.88 & 50.52 & 47.82 \\
        CRIS~\cite{cris} & CLIP-ResNet-101 & 56.27 & 55.34 & - & 63.42 & 63.82 & - & 51.79 & 51.04  & - \\
        LAVT~\cite{lavt} & Swin-B & 58.40 & 57.64 & 49.32 & 65.90 & 65.32 & 49.25 & 55.83 & 55.04 & 48.46 \\
        ReLA~\cite{rela} & Swin-B & 63.60 & 62.42 & 56.37 & 70.03 & 69.26 & 59.02 & 61.02 & 59.88 & 58.40 \\
        EEVG~\cite{eevg} & ViT-B & 62.75 & 64.04 & - & 70.93 & 71.65 & - & 62.79 & 62.77  & - \\
        HDC~\cite{hdc} & Swin-B & 68.28 & 65.42 & \underline{63.38} & \underline{72.52} & \underline{71.60} & 65.29 & 63.85 & 62.79 & 60.68 \\
        \hline
        \multicolumn{11}{l}{\textbf{Proposal-based Methods:}} \\
        MattNet~\cite{mattnet} & ResNet-101 & 48.24 & 47.51 & 41.15 & 59.30 & 58.66 & 44.04 & 46.14 & 45.33 & 41.32 \\
        \rowcolor{aliceblue!90} 
        \textbf{PropVG} & BEiT3-ViT-B & \textbf{73.29} & \textbf{69.23} & \textbf{72.83} & \textbf{74.43} & \textbf{74.20} & \textbf{69.87} & \textbf{65.87} & \textbf{64.76} & \textbf{64.97} \\

        \specialrule{.1em}{.05em}{.05em}
        \end{tabular}
    }
    \vspace{-5pt}
    \caption{Comparison with the SOTA methods on gRefCOCO (GRES)~\cite{rela}. \textbf{-QW-7B} means Qwen-VL-7B-Chat~\cite{qwen-vl}.}
    \vspace{-10pt}
    \label{tab:sota_gres}
\end{table*}

\begin{table*}[t]
        \begin{minipage}{0.58\textwidth}
        \makeatletter\def\@captype{table}
        \vspace{-6pt}
        \centering
        \footnotesize
        \setlength{\tabcolsep}{4.0pt}
        \renewcommand\arraystretch{1.2}
        \resizebox{1.0\linewidth}{!}{
            \begin{tabular}{l|ccc|ccc|ccc}
                \specialrule{.1em}{.05em}{.05em}
                \multirow{2}{*}{Method} & \multicolumn{3}{c|}{R-RefCOCO}  & \multicolumn{3}{c|}{R-RefCOCO+} & \multicolumn{3}{c}{R-RefCOCOg} \\
                \cline{2-10}
                & mIoU & mRR & rIoU & mIoU & mRR & rIoU & mIoU & mRR & rIoU  \\
                \hline
                CRIS~\cite{cris}  & 43.58 & 76.62 & 29.01 & 32.13 & 72.67 & 21.42 & 27.82	& 74.47 &	14.60 \\
                EFN~\cite{efn}    & 58.33  & 64.64 & 32.53 & 37.74  & 77.12 & 24.24 & 32.53 & 75.33 & 19.44 \\
                VLT~\cite{vlt}    &  61.66  & 63.36 & 34.05 & 50.15 & 75.37 & 34.19 & 49.67 & 67.31 & 31.64\\
                LAVT~\cite{lavt}   & 69.59 & 58.25 & 36.20 & 56.99 & 73.45 & 36.98 & 59.52 &  61.60 & 34.91 \\
                LAVT+~\cite{lavt}     & 54.70 & 82.39 & 40.11 & 45.99 & 86.35 & 39.71 & 47.22 & 81.45 & 35.46 \\
                RefSegformer~\cite{rris} & 68.78 & 73.73 & 46.08 & 55.82 & 81.23 & 42.14 & 54.99 &  71.31 & 37.65 \\ 
                HDC~\cite{hdc}& \underline{74.35} & \underline{83.69} & \underline{52.81} & \underline{64.85} & \underline{87.51} & \underline{49.09} & \underline{65.11} & \underline{84.19} & \underline{43.85} \\
                \hline
                \rowcolor{aliceblue!90} 
                \textbf{PropVG} & \textbf{75.86} & \textbf{92.39} & \textbf{62.34} & \textbf{69.39} & \textbf{94.48} & \textbf{59.04} & \textbf{69.20} & \textbf{92.88} & \textbf{55.09} \\
                \specialrule{.1em}{.05em}{.05em} 
            \end{tabular}
        }
        \vspace{-5pt}
        \caption{Comparison with SOTA methods on the R-RefCOCO/+/g~\cite{rris} dataset.}
        \label{table:sota_rrefcoco}
        \vspace{-10pt}
        \end{minipage}
        \hspace{0.02\textwidth}
        %
        \begin{minipage}{0.38\textwidth}
        \makeatletter\def\@captype{table}
        \centering
        \footnotesize
        \renewcommand\arraystretch{1.2}
        \setlength{\tabcolsep}{4.0pt}
        \resizebox{0.9\linewidth}{!}{
            \begin{tabular}{l|c|ccc}
                \specialrule{.1em}{.05em}{.05em}
                {Method}& {Backbone} & oIoU & mIoU & Acc. \\ 
                \hline
                \rowcolor{ggray!20}
                \multicolumn{5}{c}{\textit{MLLM Methods}} \\
                \hline
                LISA-V-7B~\cite{lisa} & SAM-ViT-H & 65.39 & 66.41 & 93.39 \\
                GSVA-V-7B~\cite{gsva} & SAM-ViT-H & 68.13 & 68.29 & 94.59\\
                \hline
                \rowcolor{ggray!20}
                \multicolumn{5}{c}{\textit{Specialist Methods}} \\
                \hline
                MCN~\cite{mcn} & DarkNet-53 & 54.70 & 55.03 & 75.81 \\ 
                VLT~\cite{vlt} & DarkNet-53 & 60.43 & 60.21 & 79.26 \\ 
                LAVT~\cite{lavt} & Swin-B & 64.78 & 64.45 & 83.11 \\ 
                DMMI~\cite{refzom} & Swin-B & 68.21 & 68.77 & 87.02 \\
                HDC~\cite{hdc} & Swin-B & \underline{69.31} & \underline{68.81} & \underline{93.34}\\
                \hline
                \rowcolor{aliceblue!90}  
                \textbf{PropVG} & BEiT3-ViT-B & \textbf{71.95} & \textbf{71.15} & \textbf{98.11}\\
                \specialrule{.1em}{.05em}{.05em}
            \end{tabular}
        }
        \vspace{-5pt}
        \caption{Comparison with SOTA methods on the Ref-ZOM~\cite{refzom} dataset.}
        \label{table:sota_ref_zom}
        \vspace{-10pt}
        \end{minipage}
\end{table*}

\begin{table}
\setlength{\tabcolsep}{2pt}
\renewcommand{\arraystretch}{1.2}
\centering
\footnotesize
\resizebox{1.0\linewidth}{!}{
    \begin{tabular}{l|c|cc|cc|cc}
        \specialrule{.1em}{.05em}{.05em}
        \multirow{2}{*}{Methods} &\multirow{2}{*}{Encoder} &  \multicolumn{2}{c|}{Val} & \multicolumn{2}{c|}{TestA} & \multicolumn{2}{c}{TestB} \\
        &  & F1score   & N-acc.  &F1score   &  N-acc.   &F1score    & N-acc.   \\
        \hline
        VLT~\cite{vlt} &DarkNet-53& 36.6 & 35.2  & 40.2 & 34.1 & 30.2 & 32.5 \\
        MDETR~\cite{mdetr} &ResNet-101& 42.7 & 36.3 & 50.0 & 34.5 & 36.5 & 31.0   \\
        UNINEXT~\cite{uninext} &ResNet-50 & 58.2  &  50.6 & 46.4 & 49.3 & 42.9 & 48.2 \\
        SimVG~\cite{simvg} & BEiT3-ViT-B & \underline{62.1} & \underline{54.7} & \underline{64.6} &\underline{57.2} & \underline{54.8}& \underline{57.2} \\
        \hline
        \rowcolor{aliceblue!90}  
        \textbf{PropVG} & BEiT3-ViT-B &  \textbf{72.2} & \textbf{72.8} & \textbf{68.8} &\textbf{69.9} & \textbf{59.0} & \textbf{65.0}\\
        \specialrule{.1em}{.05em}{.05em}
    \end{tabular}
}
\vspace{-5pt}
\caption{GREC benchmark results on the gRefCOCO~\cite{grec} dataset. The threshold is set to 0.7 for all the methods.}
\vspace{-5pt}
\label{table:sota_grec}
\end{table}

\subsubsection{Classic Visual Grounding}
\label{subsubsec:classic_visual_grounding}
In this section, we compare the performance of PropVG against various SOTA models on the REC and RES tasks. As shown in Table~\ref{table:sotaonpretrain}, we categorize existing methods into two main classes: MLLM and Specialist methods. The Specialist methods are further divided into Direct Referring methods (including one-stage and transformer-based) and Proposal-based methods (two-stage). 

\noindent{\bf Results on REC.}
We begin by comparing PropVG with existing methods on the RefCOCO/+/g (REC)~\cite{refcoco/+,refcocog} dataset. Compared to classic proposal-based model MattNet~\cite{mattnet}, we achieve substantial improvements not only in performance (+14\% on average) but also in inference efficiency ($4\times$ faster). Moreover, even when juxtaposed with direct referring methods that do not necessitate proposals, our method surpasses the recently established SOTA model OneRef~\cite{oneref}, which also employs BEiT-3 architecture. Specifically, we attain an average improvement of 0.5\% on RefCOCO, 1.2\% on RefCOCO+, and 0.9\% on RefCOCOg dataset. Furthermore, our approach demonstrates strong competitiveness when compared to recent MLLM-based methods, while utilizing significantly less parameters to achieve these results ($\approx$ 0.2B v.s. 7B+).

\noindent{\bf Results on RES.}
We further evaluate our method on the RefCOCO/+/g (RES)~\cite{refcoco/+,refcocog} dataset. Compared to EEVG~\cite{eevg}, which also adopts a ViT-B backbone, our method achieves an average mIoU improvement of 1.4\%, 4.0\%, and 2.4\% on RefCOCO/+/g datasets, respectively. To ensure comprehensive evaluation, we additionally report the oIoU metric in the last line of Table~\ref{table:sotaonpretrain}.

\subsubsection{Generalized Visual Grounding}
\label{subsubsec:generalized_visual_grounding}
To evaluate the effectiveness of our PropVG in the generalized setting, we conduct a comparative analysis with existing methods on the gRefCOCO(GREC~\cite{grec}/GRES~\cite{rela}), Ref-ZOM~\cite{refzom} and R-RefCOCO/+/g~\cite{rris} datasets. 

\noindent{\bf Results on GRES.}  
As shown in Table~\ref{tab:sota_gres}, the results demonstrate that our method achieves superior performance across all the metrics on the three evaluation sets of the large-scale gRefCOCO~\cite{rela} benchmark. Specifically, compared to the SOTA method HDC~\cite{hdc}, PropVG shows significant improvements of +5.0\%, +1.9\%, and +2.0\% in gIoU on the val, testA, and testB sets, respectively. 
Furthermore, we report our results on the Ref-ZOM benchmark~\cite{refzom} in Table~\ref{table:sota_ref_zom}. Our method consistently outperforms other approaches, achieving improvements of 4.8\% in Accuracy, 2.6\% in oIoU, and 2.2\% in mIoU. Notably, our approach surpasses MLLM-based models~\cite{lisa,gsva,sam4mllm}, which uses the capabilities of both SAM~\cite{sam} and LLM~\cite{llama}. 
Additionally, we extend our evaluation to the R-RefCOCO/+/g datasets~\cite{rris}. As shown in Table~\ref{table:sota_rrefcoco}, our method achieves substantial improvements of 9.5\%, 10.0\%, and 11.2\% in rIoU for the R-RefCOCO/+/g tasks compared to HDC~\cite{hdc}.

\noindent{\bf Results on GREC.}  
Beyond general segmentation, our PropVG model is also equipped to handle detection tasks. We evaluate the detection performance of PropVG on the gRefCOCO~\cite{grec} dataset, comparing it with existing SOTA methods. The results, as presented in Table~\ref{table:sota_grec}, highlight that, under the same threshold of 0.7, PropVG significantly outperforms the current SOTA method SimVG~\cite{simvg}, with improvements of +10.1\%, +4.2\%, and +4.2\% in F1score on the validation, testA, and testB sets, respectively.

\subsection{Ablation Studies}
\label{subsec:ablation}

\begin{table}
    \centering
    \footnotesize
    \setlength{\tabcolsep}{2.0pt}
    \renewcommand\arraystretch{1.2}
    \resizebox{0.85\linewidth}{!}{
    \begin{tabular}{l|cccc}
    \specialrule{.1em}{.05em}{.05em}
    Method & F1score & N-acc. & gIoU & cIoU \\
    \hline 
    Basic Setting & 63.41 & 64.11 & 65.98 & 62.43 \\
    \hline
    + SimFPN & 65.14 & 68.02 & 66.86 & 63.91 \\
    + UNet Decoder & 65.87 & 69.19 & 68.16 & 65.30 \\
    + Multi-scale Deformable Decoder & 67.44 & 69.47 & 69.10 & 65.77 \\
    + Channel Splitting & 67.98 & \textbf{70.44} & 69.59 & 66.22 \\
    \rowcolor{aliceblue!90}  + Query Proj. (Baseline) & \textbf{68.81} & 70.39 & \textbf{69.85} & \textbf{66.24}\\
    \rowcolor{gray!10} - Foreground Supervision & 66.83 & 61.06 & 66.37 & 64.99 \\
    \specialrule{.1em}{.05em}{.05em}
    \end{tabular}
    }
    \vspace{-5pt}
    \caption{Basic setting improvements.}
    \vspace{-10pt}
    \label{tab:ab_basic_setting}
\end{table}

\subsubsection{Basic Setting Improvements}
\label{subsec:basic_setting_improvements}

To enhance the foundational performance of the baseline model, we introduce several key components. The basic model directly utilizes the visual features $f_v$ in the segmentation branch to predict segmentation results through a SegHead, while the detection branch employs the original DETR~\cite{detr} decoder structure. 
Building upon this foundation, we integrate several design elements, including SimFPN~\cite{vitdet}, UNet Decoder~\cite{unet}, Multi-scale Deformable Decoder~\cite{deformabledetr}, Channel Splitting, and Query Proj., to improve the baseline model's performance, as shown in Table~\ref{tab:ab_basic_setting}. SimFPN extends the single-scale features of ViT into hierarchical representations, enhancing the model's fine-grained capabilities for both detection and segmentation. The UNet Decoder addresses the segmentation task’s need for multi-scale spatial information. The Multi-scale Deformable Decoder introduces multi-scale deformable attention, effectively capturing the shape information of targets. Channel Splitting divides feature channels into two parts, each dedicated to detection and segmentation tasks, improving task-specific adaptation. The Query Proj. enhances $Q_{prop}$ for better alignment with the referring task, enriching the representation transfer from detection to referring tasks.
As a result, our enhanced baseline model demonstrates substantial improvements over the basic setting, with +5.4\% in F1score, +6.3\% in N-acc., +3.9\% in gIoU.

Additionally, when foreground supervision is omitted during the proposal stage and the matching process is substituted with the original direct referring structure, we observe an average performance decline of 2\%. This reduction in performance can be attributed to the reformulation of the VG task in PropVG. Specifically, the proposal-based methodology redefines the task as a two-step process: candidate generation followed by binary classification. This transformation effectively lowers the training complexity compared to the conventional direct referring approach.
Moreover, we contend that the incorporation of additional spatial supervision significantly bolsters the model’s capacity to capture global object information. This enhancement in spatial awareness contributes to a more robust and comprehensive understanding of the target, ultimately elevating the model’s overall performance. More explanations are provided in the Appendix~F.

\begin{table*}[t]
        \begin{minipage}{0.31\textwidth}
        \makeatletter\def\@captype{table}
        \centering
        \footnotesize
        \setlength{\tabcolsep}{2.0pt}
        \renewcommand\arraystretch{1.2}
        \resizebox{0.9\linewidth}{!}{
        \begin{tabular}{cc|cccc}
        \specialrule{.1em}{.05em}{.05em} 
        CRS & MTD & F1score & N-acc. & gIoU & cIoU \\
        \hline
                   &            &  68.81 & 70.39 & 69.85 & 66.24 \\
        \checkmark &            &  70.61 & 74.78 & 71.30 & 66.91 \\
                   & \checkmark &  70.28 & 73.17 & 71.26 & 66.66 \\
        \rowcolor{aliceblue!90}  \checkmark & \checkmark & \textbf{71.00} & \textbf{75.81} & \textbf{71.99} & \textbf{67.23} \\
        \specialrule{.1em}{.05em}{.05em}
        \end{tabular}
        }
        \vspace{-5pt}
        \caption{Effectiveness of the core modules.}
        \label{tab:ab_module_effectiveness}
        \vspace{-5pt}
        \end{minipage}
        \hspace{0.005\textwidth}
        %
        \begin{minipage}{0.33\textwidth}
        \makeatletter\def\@captype{table}
        \centering
        \footnotesize
        \renewcommand\arraystretch{1.2}
        \setlength{\tabcolsep}{2.0pt}
        \resizebox{0.9\linewidth}{!}{
        \begin{tabular}{l|cccc}
        \specialrule{.1em}{.05em}{.05em}
        Method & F1score & N-acc. & gIoU & cIoU \\
        \hline
        Baseline & 68.81 & 70.39 & 69.85 & 66.24\\
        Sent. Con. & 69.70 & 73.39 & 70.84 & 66.44\\
        Word Con. & 70.12 & 73.88 & 70.93 & 66.25 \\
        Sent.-Word Con. & 70.18 & 74.11 & 71.05 & 66.54 \\
        \rowcolor{aliceblue!90}  CRS & \textbf{70.61} & \textbf{74.78} & \textbf{71.30} & \textbf{66.91} \\
        \specialrule{.1em}{.05em}{.05em}
        \end{tabular}
        }
        \vspace{-5pt}
        \caption{Ablation on CRS module.}
        \label{tab:ab_CRS}
        \vspace{-5pt}
        \end{minipage}
        \hspace{0.005\textwidth}
        \begin{minipage}{0.34\textwidth}
        \makeatletter\def\@captype{table}
        \centering
        \footnotesize
        \setlength{\tabcolsep}{2.0pt}
        \renewcommand\arraystretch{1.2}
        \resizebox{0.9\linewidth}{!}{
        \begin{tabular}{l|cccc}
        \specialrule{.1em}{.05em}{.05em}
        Method & F1score & N-acc. & gIoU & cIoU \\
        \hline 
        Baseline & 68.81 & 70.39 & 69.85 & 66.24 \\
        Query SPCA & 69.54 & 71.02 & 70.03 & 66.13 \\
        Mask SPCA & 69.13 & 71.87 & 70.87 & 66.30 \\
        Query-Mask SPCA & 69.88 & 72.35 & 71.01 & 66.43 \\
        \rowcolor{aliceblue!90}  MTD & \textbf{70.28} & \textbf{73.17} & \textbf{71.26} & \textbf{66.66} \\
        \specialrule{.1em}{.05em}{.05em}
        \end{tabular}
        }
        \vspace{-5pt}
        \caption{Ablation on MTD module.}
        \label{tab:ab_MTD}
        \vspace{-5pt}
        \end{minipage}
\end{table*}

\begin{figure*}
    \centering
    \includegraphics[width=1.0\linewidth]{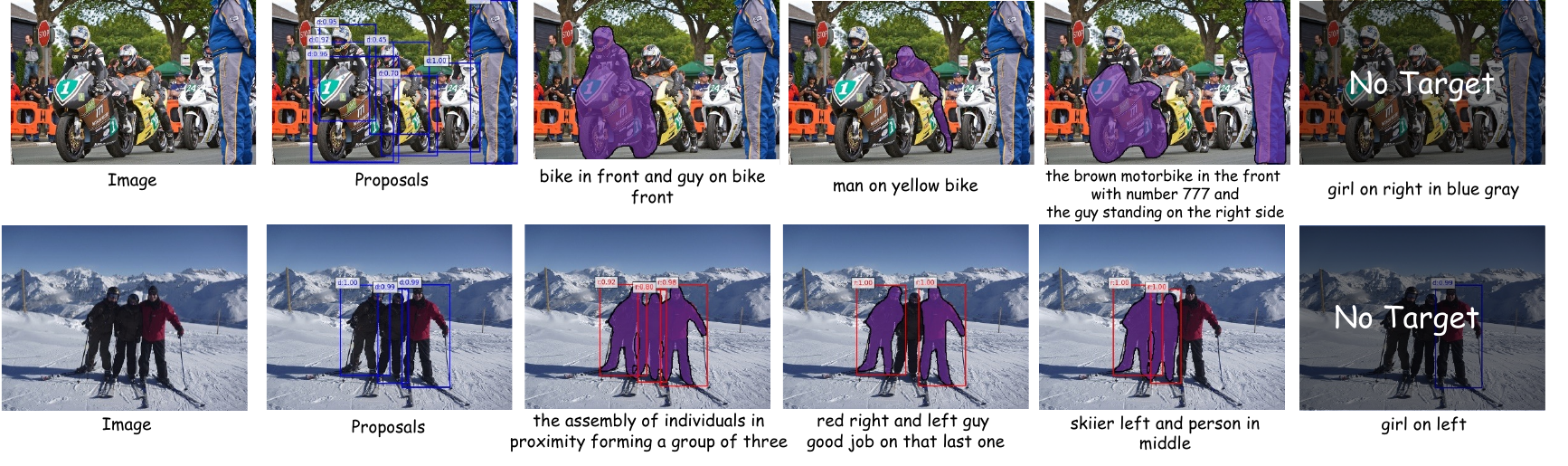}
    \vspace{-15pt}
    \caption{Visualization of predicted proposals, alongside the detection and segmentation results under various textual conditions.}
    \label{fig:visualize_final}
    \vspace{-5pt}
\end{figure*}

\subsubsection{Module Effectiveness}  
\label{subsubsec:module_effectiveness}
Table~\ref{tab:ab_module_effectiveness} presents the results of the core modules. The CRS module enhances the referential discrimination of the query through sentence- and word-level contrastive learning, resulting in +1.4\% in F1score and +1.5\% in gIoU. The MTD module improves the stability of target existence prediction by integrating object- and semantic-level features, and by combining refer and pixel scores, leading to +1.5\% in F1score and +1.4\% in gIoU. 
The combination of both modules further boosts the performance, with F1score and N-acc. increased by 2.2\% and 5.4\%, respectively, and gIoU and cIoU improved by 2.1\% and 1.0\%, respectively.

\subsubsection{Analysis of the CRS module}
\label{subsubsec:crs_analysis}
The CRS module computes the referring score for queries by interacting with textual features. It follows the contrastive learning paradigm, using adaptive weighting to combine sentence- and word-level query-to-text similarities, thereby enhancing the discriminability of the referring score. The experimental results of the CRS module are presented in Table~\ref{tab:ab_CRS}. 
We use a linear layer as the baseline. `Sent. Con.' refers to using only sentence-level contrastive learning, while `Word Con.' denotes word-level contrastive learning. `Sent.-Word Add.' represents averaging the results of both sentence- and word-level contrastive learning. `CRS' corresponds to the proposed contrastive-based refer scoring module.
The results demonstrate that, compared to the baseline, both sentence- and word-level contrastive learning independently enhance performance, validating that introducing contrastive learning with textual features strengthens the query’s referential discrimination. Additionally, combining both sentence- and word-level contrastive learning further improves performance, confirming their complementarity. 
Finally, our CRS achieves notable improvements, with F1score and N-acc. increased by 1.8\% and 3.4\%, respectively, and gIoU and cIoU improved by 1.5\% and 0.7\%, respectively.

\subsubsection{Analysis of the MTD module}
\label{subsubsec:mtd_analysis}

The MTD module enhances the stability of target existence prediction by incorporating object- and semantic-level features, combined with refer and pixel scores. The experimental results are shown in Table~\ref{tab:ab_MTD}.
SPCA is presented in Eq.~\ref{eq:spca}. `Query SPCA' indicates that $Q_{exist}$ interacts only with $Q_{prop}^{'}$, while `Mask SPCA' indicates that $Q_{exist}$ interacts only with semantic feature $F_s$. `Query-Mask SPCA' sequentially interacts $Q_{exist}$ with both query and mask. The `MTD' module further weights the results of $S_{ref}$ and $M_{seg}$ to enhance performance.
Experimental results demonstrate that `Query SPCA' and `Mask SPCA' improve the F1score and gIoU by 0.7\% and 1.0\%, respectively. Query-Mask SPCA further boosts performance. Finally, the MTD module improves the F1score and N-acc. by 1.5\% and 2.8\%, respectively, with gIoU and cIoU improving by 1.4\% and 0.4\%. These results indicate that the introduction of the MTD module helps the model analyze global target information from both the object- and semantic-level perspectives, enhancing the stability of target existence prediction.

\subsection{Qualitative Analysis}
\label{subsec:visualization}

Fig.~\ref{fig:visualize_final} illustrates PropVG’s foreground extraction capability, along with the corresponding referred targets and segmentation masks under different textual conditions. Additional visualizations are provided in Appendix~G. Moreover, we conduct a visualization analysis of the relationship between proposal generation and text, with specific results presented in Appendix~F.2.

\section{Conclusion}
\label{sec:conclusion}
In this paper, we introduce \textbf{PropVG}, a novel end-to-end multi-task collaborative framework for generalized visual grounding. Unlike prior two-stage structures, PropVG is the first proposal-based approach to eschew dependence on pre-trained object detectors, thereby mitigating the suboptimal performance and computational inefficiencies prevalent in prior methods.
To rigorously assess the alignment between region proposals and referring expressions, we propose a contrastive-based referent scoring module. This module dynamically captures the relationships between proposals and expressions by integrating perspectives at both the sentence and word levels. Additionally, we introduce a multi-granularity target discrimination module, which bolsters the reliability of target existence prediction by leveraging the interaction of multi-granularity information in generalized scenarios.
Experimental results demonstrate that PropVG achieves SOTA performance across ten datasets.

\section*{Acknowledgements}
This work is supported by the National Natural Science Foundation of China under Nos. 62276061 and 62436002.
This work is also supported by Research Fund for Advanced Ocean Institute of Southeast University (Major Program MP202404). This work is also supported by the SEU Innovation Capability Enhancement Plan for Doctoral Students (\text{CXJH\_SEU 25125}).

{
    \small
    \bibliographystyle{ieeenat_fullname}
    \bibliography{main}
}

\clearpage
\setcounter{page}{1}
\setcounter{section}{0}
\maketitlesupplementary
\renewcommand{\thesection}{\Alph{section}} 

\section{Datasets}
\label{appendix:dataset}
\paragraph{RefCOCO/RefCOCO+~\cite{refcoco/+}} are collected using a two-player game. RefCOCO has 142,209 annotated expressions for 50,000 objects in 19,994 images, and RefCOCO+ consists of 141,564 expressions for 49,856 objects in 19,992 images. These two datasets are split into training, validation, test A and test B sets, where test A contains images of multiple people and test B contains images of multiple instances of all other objects. Compared to RefCOCO, location words are banned from the referring expressions in RefCOCO+, which makes it more challenging.
\paragraph{RefCOCOg~\cite{refcocog}} is collected on Amazon Mechanical Turk, where workers are asked to write natural language referring expressions for objects. RefCOCOg consists of 85,474 referring expressions for 54,822 objects in 26,711 images. RefCOCOg has longer, more complex expressions (8.4 words on average), while the expressions in RefCOCO and RefCOCO+ are more succinct (3.5 words on average), which makes RefCOCOg particularly challenging. We use the UMD partition for RefCOCOg as it provides both validation and testing sets and there is no overlapping between training and validation images.
\paragraph{gRefCOCO~\cite{rela}} comprises 278,232 expressions, including 80,022 referring to multiple targets and 32,202 to empty targets. It features 60,287 distinct instances across 19,994 images, which are divided into four subsets: training, validation, testA, and testB, following the UNC partition of RefCOCO.
\paragraph{Ref-ZOM~\cite{refzom}} is derived from the COCO dataset, consisting of 55,078 images and 74,942 annotated objects. Of these, 43,749 images and 58,356 objects are used for training, while 11,329 images and 16,586 objects are designated for testing. Annotations cover three scenarios: one-to-zero, one-to-one, and one-to-many, corresponding to empty-target, single-target, and multiple-target cases in GRES, respectively.
\paragraph{R-RefCOCO~\cite{rris}} includes three variants: R-RefCOCO, R-RefCOCO+, and R-RefCOCOg, all based on the classic RES benchmark, RefCOCO+/g. Only the validation set adheres to the UNC partition principle, which is officially recognized for evaluation. The dataset formulation incorporates negative sentences into the training set at a 1:1 ratio with positive sentences.

\section{Metrics} 
\label{appendix:metrics}

For GRES~\cite{rela}, we evaluate our model using gIoU, cIoU, and N-acc metrics. For Ref-ZOM~\cite{refzom}, we use oIoU and mIoU. R-RefCOCO~\cite{rris} utilizes mIoU, mRR, and rIoU metrics, as defined in their respective benchmarks. The gIoU is calculated by averaging the IoU across all instances for each image, assigning a value of 1 to true positives in cases of empty targets and 0 to false negatives. The cIoU metric measures the ratio of intersection pixels to union pixels. In Ref-ZOM, mIoU calculates the average IoU for images containing referred objects, while oIoU corresponds to cIoU. For R-RefCOCO, rIoU evaluates segmentation quality, incorporating negative sentences and assigning equal weight to positive instances in the mIoU calculation. N-acc. in gRefCOCO and Acc. in Ref-ZOM both represent the ratio of correctly classified empty-target expressions to the total number of empty-target expressions. Additionally, mRR in R-RefCOCO computes the recognition rate for empty-target expressions, averaged across the dataset.

For GREC~\cite{grec}, we assess the percentage of samples with an F1score of 1, using an IoU threshold of 0.5. A predicted bounding box is considered a true positive (TP) if it overlaps with a ground-truth box with an IoU of at least 0.5. If multiple predictions match, only the one with the highest IoU is counted as TP. Unmatched ground-truth boxes are false negatives (FN), and unmatched predictions are false positives (FP). The F1score for each sample is computed as $\text{F1score} = \frac{2TP}{2TP+FN+FP}$, with a score of 1 indicating a successful prediction. For samples without targets, the F1score is 1 if no predictions are made; otherwise, it is 0.

For REC, we evaluate accuracy based on the grounding results. A predicted region is considered correct if the IoU with the ground truth exceeds 0.5. For RES, we employ mean Intersection over Union (mIoU) between predicted masks and ground truth as the evaluation metric.

\section{Additional Methods}
\label{appendix:method}

\subsection{Dataset Construction}
\label{appendix:method_dataset}

To enrich the existing referential datasets with information on foreground objects, we retrieve all corresponding foreground targets from the original COCO dataset based on the \texttt{image\_id} present in the datasets. Unlike traditional general object detection tasks, our approach focuses solely on prominent foreground objects while minimizing the emphasis on weaker or occluded instances. 
During the dataset construction, we filter out targets in the original COCO dataset where \texttt{is\_crowd=1}. Additionally, we exclude objects with an absolute area smaller than 100 pixels and retain only those with a relative area greater than 0.05 and less than 0.8. This selection process ensures that the objects in our dataset consist exclusively of significant foreground targets. 
The primary motivation behind these choices is to accelerate the model's convergence while directing attention away from fine-grained small objects. Instead, we aim to enhance the model's focus on the understanding of referential relationships between text and images.

\subsection{Post-process}
\label{appendix:method_postprocess}
\begin{figure}
    \centering
    \includegraphics[width=0.9\linewidth]{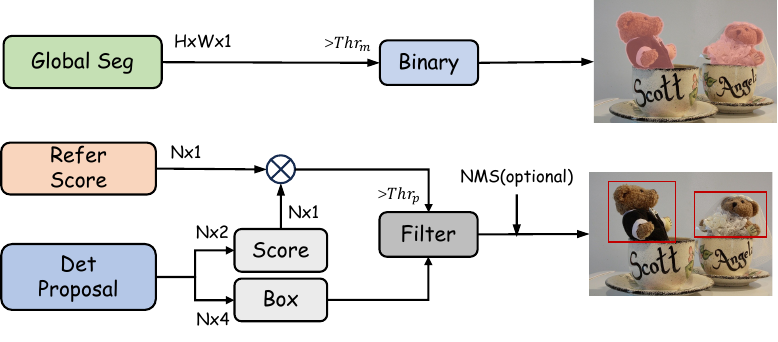}
    \caption{\textbf{Post-Process Flowchart.} We combine the Refer Score and Det Score to form the Proposal Referential Score, which is used to filter the referred target via a threshold $Thr_p$. NMS is optional and not mandatory for DETR-based architectures.}
    \label{fig:postprocess}
\end{figure}

\noindent{\textbf{Post-Processing Procedure.}} As illustrated in Fig.~\ref{fig:postprocess}, the post-processing consists of three components: Global Segmentation, Refer Score, and Det Proposal. The Det Proposal branch further decomposes into Det Score and Det Box. For the segmentation branch, we directly apply a threshold $Thr_m$ to binarize each pixel. In the target referential part, we combine the Refer Score with the Det Score, as a valid referred target should possess both high detection confidence and high referential confidence. We filter the combined score using a threshold $Thr_p$. NMS is optional in this process and not mandatory for DETR-based architectures.

\section{Additional Implementation Details}
\label{appendix:implementation}
For GVG tasks (GREC, GRES), images are resized to $320 \times 320$, and models are trained for 12 epochs.
For CVG tasks (REC, RES), images are resized to $384 \times 384$, with training for 30 epochs.
We use a unified batch size of 16 and adopt the Adam optimizer.
All experiments are conducted on four NVIDIA 4090 GPUs, without using Exponential Moving Average (EMA).
The initial learning rate is $5 \times 10^{-5}$ for the multi-modality encoder and $5 \times 10^{-4}$ for other parameters.
The learning rate decays by a factor of 0.1 at the 7th and 11th epochs.
All ablation studies are conducted at a resolution of $224 \times 224$ and trained for 10 epochs.

\section{Additional Ablation Studies}
\label{appendix:ablation_studies}

\subsection{Impact of the Hyperparameter $K$ in TAS}
\label{appendix:ablation_tas}

We analyze the effect of different $K$ values on key metrics, including N-acc., T-acc., F1score, and gIoU. The choice of $K$ directly influences the performance of the model by controlling the number of top-scored segmentation pixels considered by $S_{exist}$. 
As shown in Table~\ref{fig:postprocess}, the value of $K$ determines the extent to which segmentation results contribute to the final target existence score. A higher $K$ suppresses the existence confidence, leading to a reduction in the overall refer score. This intuitively improves N-acc. but decreases T-acc. 
A crucial aspect to discuss is the optimal number of segmentation pixels to use for weighting the existence score to maximize improvements in F1score and gIoU. Experimental results indicate that when $K = 250$, the framework achieves peak performance across these metrics.

\begin{figure*}
    \centering
    \includegraphics[width=0.9\linewidth]{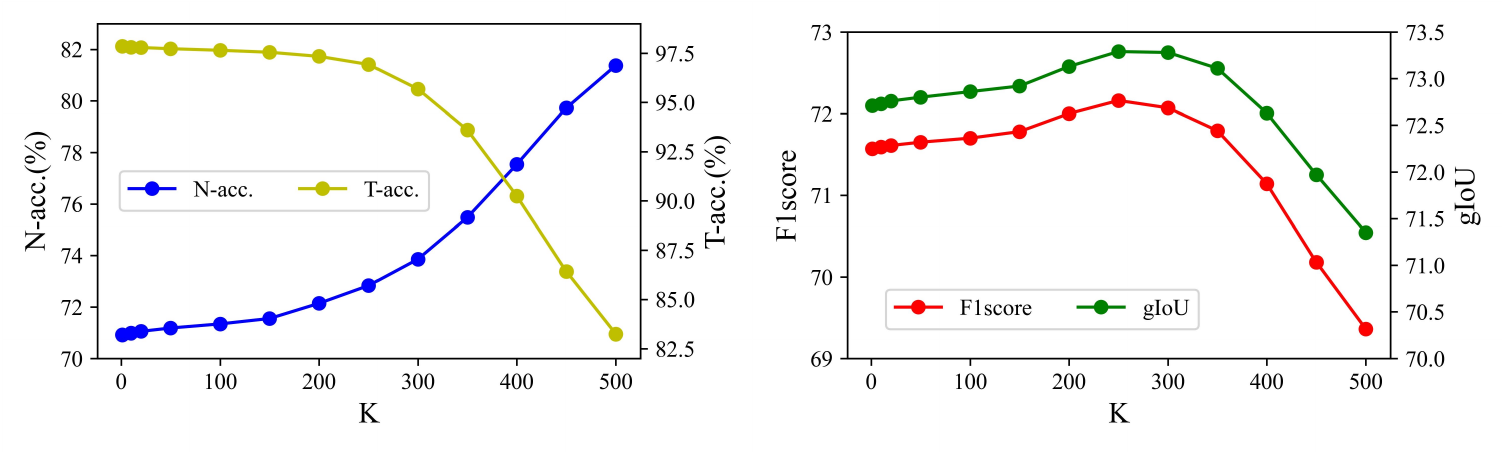}
    \vspace{-10pt}
    \caption{Impact of the Hyperparameter $K$ in TopK Average Scoring (TAS).}
    \label{fig:appendix_ab_K}
   \vspace{-5pt}
\end{figure*}

\subsection{Impact of the Other Hyperparameter}
\label{appendix:ablation_hyperparameter}
Table~\ref{tab:hyperparameter} analyzes the impact of varying individual loss weights. First, increasing the weight of $\mathcal{L}_{det}$ (ID 5) improves F1score (69.6) but slightly reduces gIoU, suggesting better detection confidence at the cost of segmentation quality. Second, tuning $\mathcal{L}_{exist}$ (ID 2 vs. ID 1) mainly affects N-acc, with higher weights enhancing classification accuracy but harming F1score. Lastly, adjusting $\mathcal{L}_{ref}$ (ID 6, 7) influences all metrics simultaneously, reflecting its central role in balancing detection, grounding, and segmentation. Overall, the default setting (ID 1) provides the best trade-off.

\begin{table}[ht]
    \centering
    \footnotesize
    \setlength{\tabcolsep}{4pt}
    \renewcommand\arraystretch{1.2}
    \resizebox{0.8\linewidth}{!}{
    \begin{tabular}{l|c|ccc}
        \specialrule{.1em}{.05em}{.05em}
        ID & $\mathcal{L}_{det}$:$\mathcal{L}_{exist}$:$\mathcal{L}_{ref}$ & F1score & N-acc  & gIoU \\
        \hline
         1 &0.1:0.2:1.0 & 69.2  & 71.0 & 69.9 \\
        \hline
         2 &0.1:0.5:1.0 & 68.4  & 71.2 & 69.5 \\
         3 &0.1:0.0:1.0 & 67.9  & 67.8 & 68.2 \\
        \hline
         4 &0.05:0.2:1.0 & 68.3 & 71.1 & 70.4 \\
         5 &0.2:0.2:1.0  & 69.6 & 71.0 & 68.0 \\
        \hline
         6 &0.1:0.2:0.5 & 69.0 & 70.3 & 68.8 \\
         7 &0.1:0.2:2.0  & 69.1 & 70.7 & 69.8 \\
        \specialrule{.1em}{.05em}{.05em}
    \end{tabular}
    }
    \vspace{-5pt}
    \caption{Ablation study on the loss weights in Eq.~\ref{eq:loss}.}
    \vspace{-5pt}
    \label{tab:hyperparameter}
\end{table}

\subsection{Impact of the Post-Process}
\label{appendix:ablation_postprocess}

In this section, we analyze two aspects: different scoring strategies and varying post-processing thresholds. For scoring strategies, we compare three methods: using the refer branch score directly, multiplying the refer and detection branch scores, and taking the average of these two scores. As shown in Table~\ref{tab:postprocess}, the direct use of the refer branch score yields the best results, as this score more accurately reflects the importance of the target.
For post-processing, we evaluate the performance under different threshold values $Thr_p$. As presented in Table~\ref{tab:abthr}, the model performs best when $Thr_p=0.9$.

\subsection{Impact of Foreground Object Filter}
\label{appendix:ablation_foreground}

\begin{figure}
    \centering
    \includegraphics[width=0.9\linewidth]{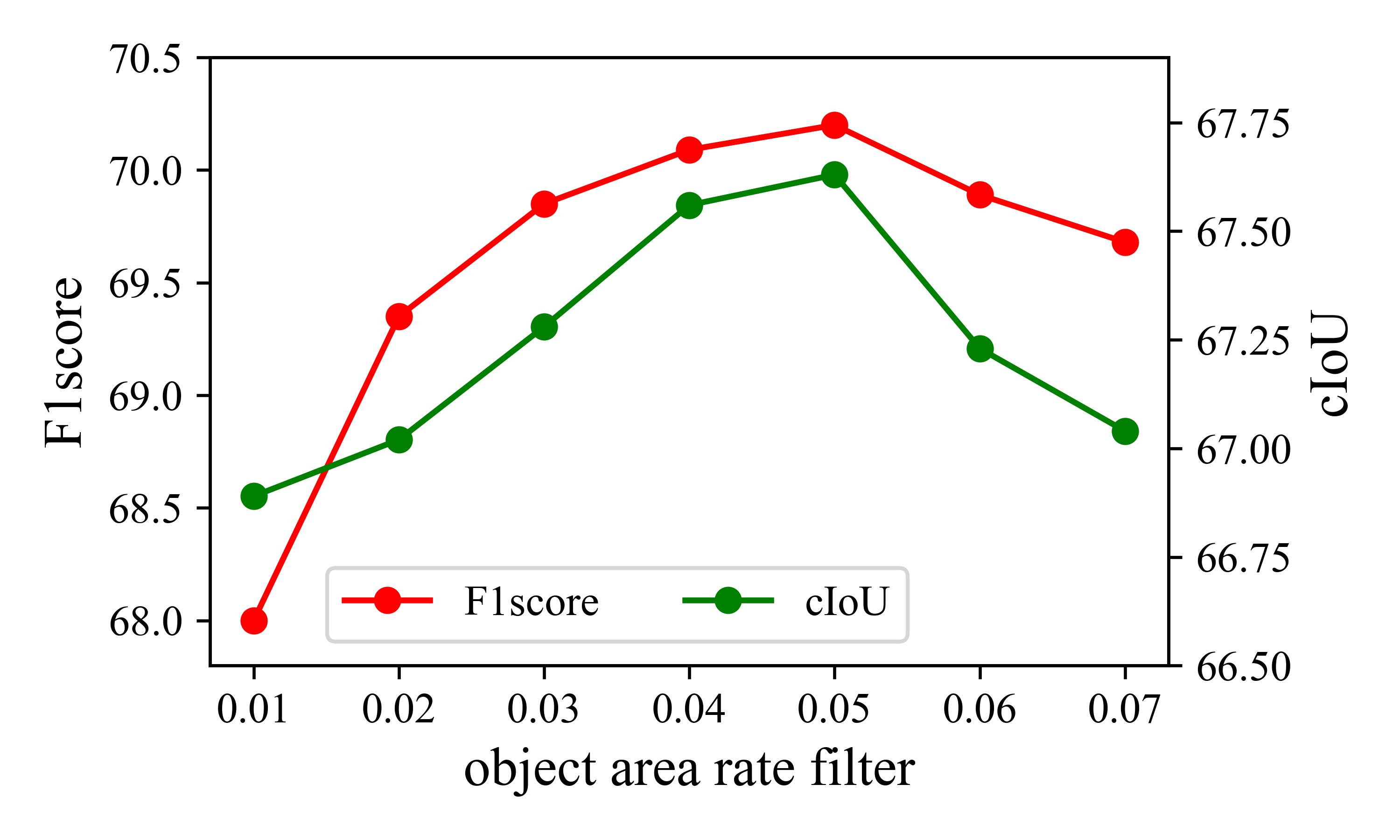}
    \vspace{-10pt}
    \caption{Impact of the lower limit on the relative area of the filtered objects.}
    \vspace{-5pt}
    \label{fig:appendix_foreground_filter}
\end{figure}

In the training process of the foreground detection branch, we establish a relative area constraint to prevent the model from overly focusing on small targets, which could weaken the optimization effects of other branches. Specifically, we set an upper limit of 0.8 for the relative area of foreground targets and a minimum absolute area of 100. We focus on examining the impact of the lower limit $R_{low}$ on model performance. As shown in Fig.~\ref{fig:appendix_foreground_filter}, setting $R_{low}$ too low introduces numerous small foreground targets into the training set, leading the model to overemphasize these weaker targets and detracting from the core referential task. Conversely, setting $R_{low}$ too high may result in the neglect of some smaller referential targets during training. Experimental validation indicates that when $R_{low}$ is set to 0.05, the model achieves optimal performance across all evaluation metrics.

\begin{table}[t]
    \centering
    \footnotesize
    \setlength{\tabcolsep}{4pt}  
    \renewcommand\arraystretch{1.2}  
    \resizebox{0.8\linewidth}{!}{
    \begin{tabular}{l|cccc}
    \specialrule{.1em}{.05em}{.05em}
    \textbf{Method} & \textbf{F1score} & \textbf{N-acc.} & \textbf{gIoU} & \textbf{cIoU} \\
    \hline
    $S_{\text{refer}}$ & 71.59 & 70.99 & 72.73 & 69.18 \\
    $S_{\text{refer}} \times S_{\text{det}}$ & 71.38 & 72.01 & 72.83 & 69.06 \\
    Avg($S_{\text{refer}}, S_{\text{det}}$) & 71.87 & 68.55 & 71.80 & 68.78 \\
    $S_{\text{refer}}$ + NMS & 71.65 & 71.34 & 72.86 & 69.20 \\
    \specialrule{.1em}{.05em}{.05em}
    \end{tabular}
    }
    \vspace{-5pt}
    \caption{Ablation study on post-processing details. The first three experiments compare different scoring strategies for target referencing: $S_{\text{refer}}$ refers to using the score from the refer branch directly; $S_{\text{refer}} \times S_{\text{det}}$ denotes the product of scores from the refer and detection branches; and Avg takes the average of the two scores. The final experiment applies Non-Maximum Suppression.}
    \vspace{-5pt}
    \label{tab:postprocess}
\end{table}

\begin{table}
    \centering
    \footnotesize
    \setlength{\tabcolsep}{2.0pt}
    \renewcommand\arraystretch{1.2}
    \resizebox{0.6\linewidth}{!}{
    \begin{tabular}{l|cccc}
    \specialrule{.1em}{.05em}{.05em}
    $Thr_{p}$ & F1score & N-acc. & gIoU & cIoU \\
    \hline
    0.5 & 65.14 & 68.02 & 66.86 & 63.91 \\
    0.6 & 65.87 & 69.19 & 68.16 & 65.30 \\
    0.7 & 67.44 & 69.47 & 69.10 & 65.77 \\
    0.8 & 67.98 & $\mathbf{70.44}$ & 69.59 & 66.22 \\
    0.9 & $\mathbf{68.81}$ & 70.39 & $\mathbf{69.85}$ & $\mathbf{66.24}$\\
    \specialrule{.1em}{.05em}{.05em}
    \end{tabular}}
    \vspace{-5pt}
    \caption{Effectiveness of $Thr_{p}$ in post-process.}
    \vspace{-5pt}
    \label{tab:abthr}
\end{table}

\begin{figure*}
    \centering
    \includegraphics[width=1.0\linewidth]{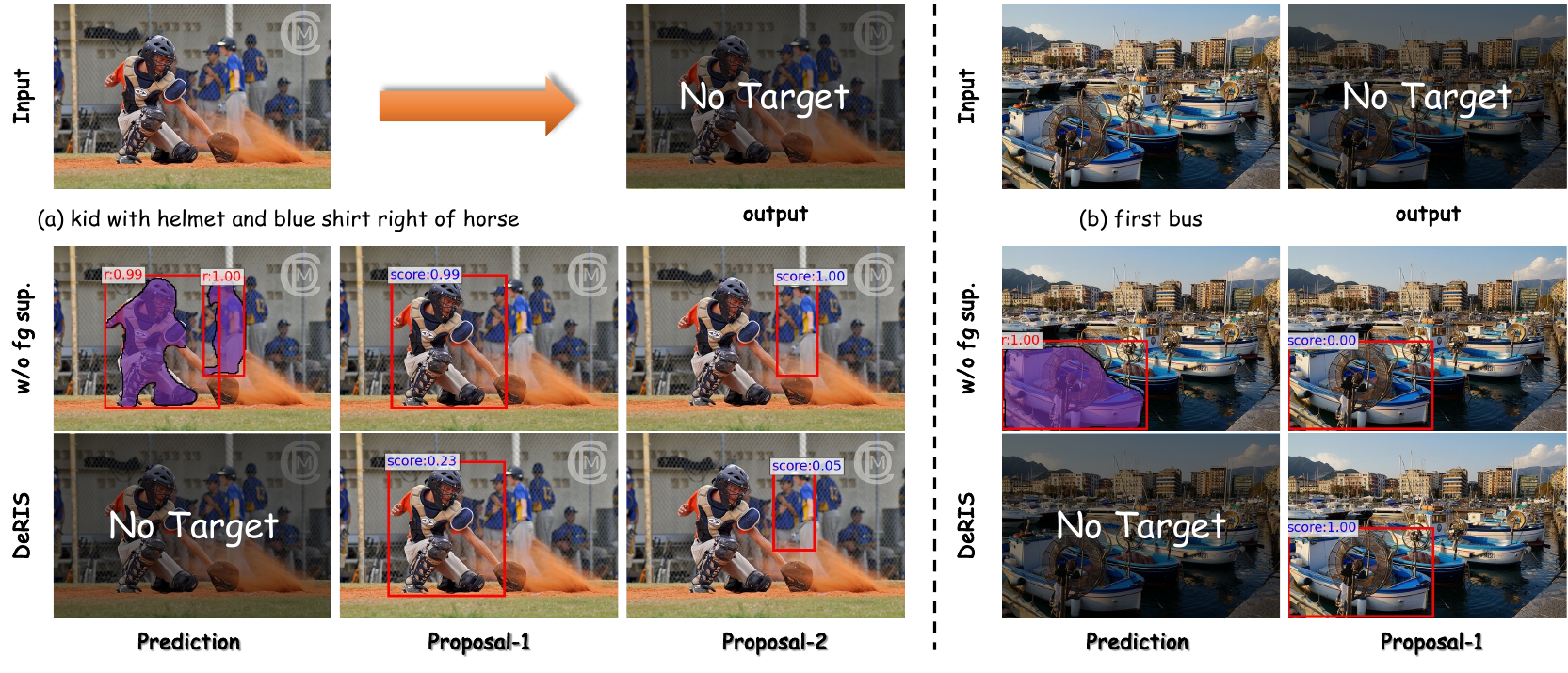}
    \vspace{-15pt}
    \caption{Illustration the effect of incorporating foreground supervision.  
    \textit{w/o fg sup.} denotes the setting where only the referred objects are used as supervision targets, potentially overlooking useful foreground cues.}
    \label{fig:foregroundsupervision}
    \vspace{-10pt}
\end{figure*}

\section{Analysis}
\label{sec:analysis}

\subsection{Foreground Supervision Analysis}
\label{appendix:foregroundsupervision}
Fig.~\ref{fig:foregroundsupervision} provides a visual comparison between two regression strategies: direct referring and our proposal-driven referring approach. In the latter, the model first generate the foreground target before subsequently discriminate the proposal's referentiality. For instance, using text (a) as a case study, the direct referring method erroneously identifies a non-referent target with high confidence. In contrast, our DeRIS effectively diminishes the confidence assigned to non-target instances, thereby reducing false positives. Experimental results indicate that the integration of foreground supervision enhances the model’s ability to differentiate between foreground and referent ones, leading to a marked reduction in false positives and a substantial improvement in generalization performance.

\subsection{Proposal Analysis}
\label{appendix:proposalanalysis}

\begin{figure*}
    \centering
    \includegraphics[width=1.0\linewidth]{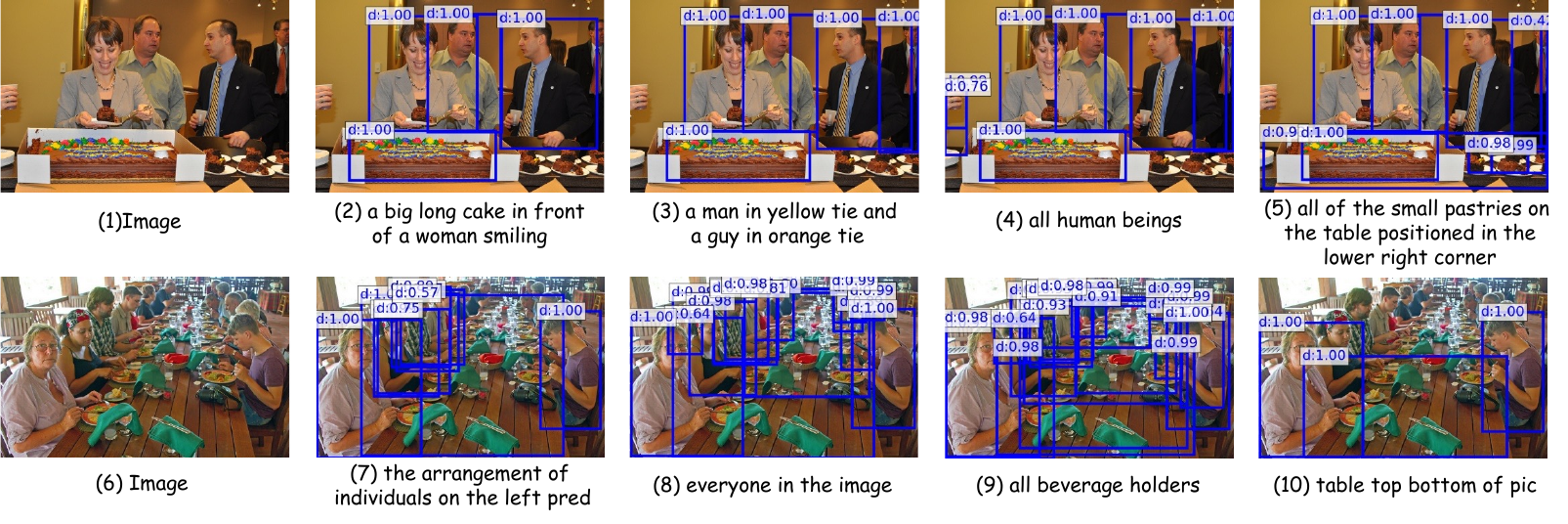}
    \vspace{-10pt}
    \caption{Illustration of foreground object prediction results under different texts.}
    \label{fig:limitation}
   \vspace{-5pt}
\end{figure*}

By examining the foreground outputs generated for the same image under varying textual descriptions, we observe that the textual input exerts a substantial influence on the generation of foreground targets. Specifically, the model prioritizes targets explicitly referenced in the text as foreground objects. For example, in Fig.~\ref{fig:limitation} (3), where the description `a guy in an orange tie' is provided, the model accurately identifies and emphasizes the corresponding target as the foreground object. In contrast, in Fig.~\ref{fig:limitation} (2), where no such description is given, the same target is not classified as foreground. Similarly, in Fig.~\ref{fig:limitation} (4), the instruction `all human beings' prompts the model to designate all humans as foreground targets, including subtle details such as the smaller hand of a person on the left, which is overlooked under alternative descriptions.

Further examples reinforce this trend. In Fig.~\ref{fig:limitation} (5), the phrase `all of the small pastries' leads the model to detect every pastry in the image as a foreground object, whereas texts omitting `pastries' exclude them from consideration. In Fig.~\ref{fig:limitation} (7), the model focuses exclusively on objects on the `left' side, as dictated by the text. In cases such as Fig.~\ref{fig:limitation} (8) and (9), where inclusive terms like `all' or `everyone' are used, the model identifies all visible targets as foreground objects. Conversely, as illustrated in Fig.~\ref{fig:limitation} (10), when the text description is misaligned with the image content, the model suppresses foreground generation entirely.

These findings underscore the pivotal role of multimodal understanding approaches, such as BEiT-3, in driving these behaviors. In such frameworks, the interaction between image and text features initiates at the encoding stage, enabling the model to concentrate attention on objects that align with the textual description when a specific reference is provided. This early-stage interplay allows textual information to directly modulate the representation of image features, amplifying responses in regions pertinent to the description while attenuating those in unrelated areas. Consequently, this mechanism facilitates precise and contextually targeted foreground generation.

\section{Additional Visualization}
\label{appendix:visualization}

\subsection{Query Visualization}
\label{appendix:query_visualization}
\begin{figure*}[t]
    \centering
    \includegraphics[width=0.9\linewidth]{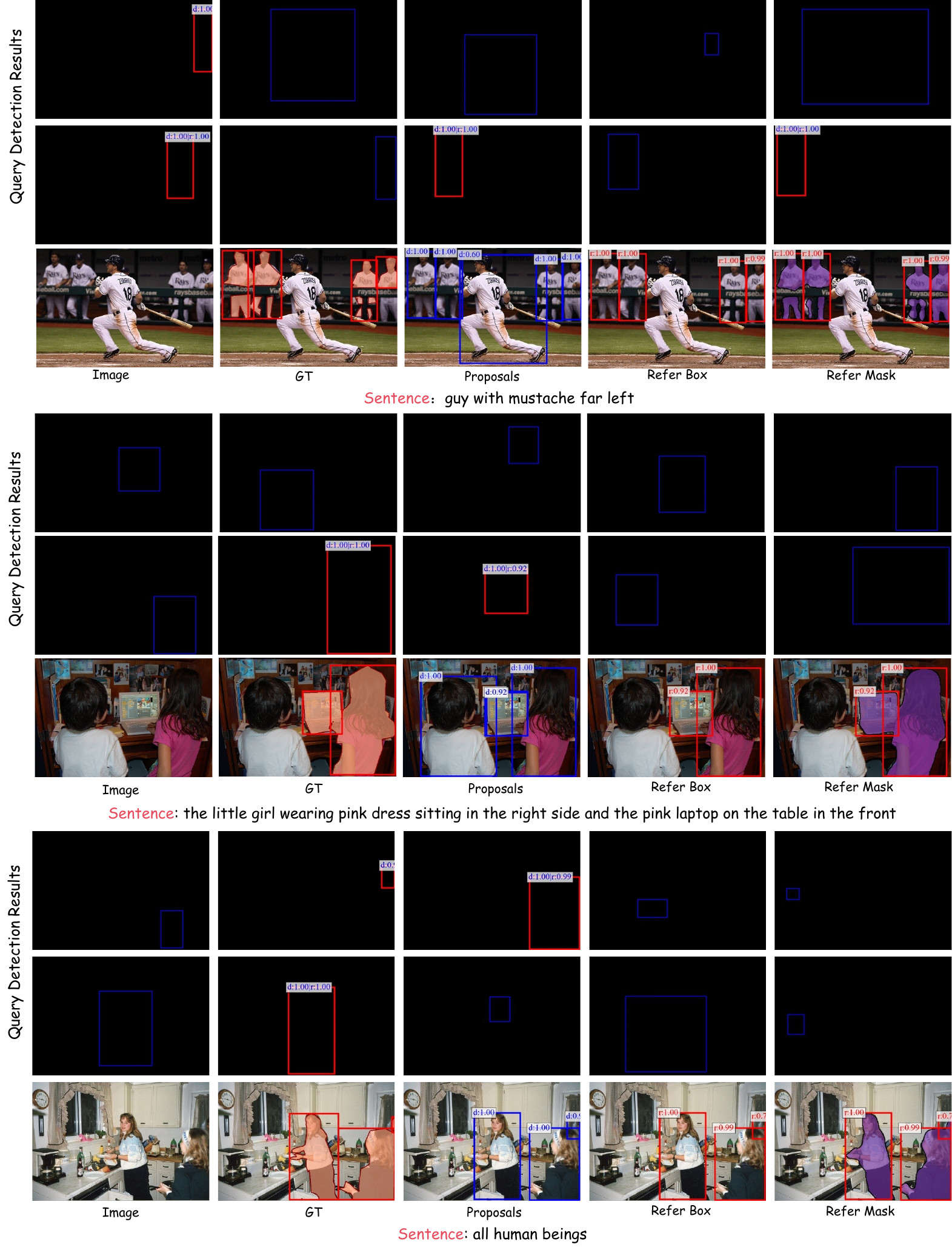}
    \vspace{-10pt}
    \caption{Visualization of object queries with corresponding detection boxes, detection scores, and referential scores. Red boxes indicate correctly referred objects, while blue boxes represent objects with low referential confidence.}
    \label{fig:appendix_visualize_query}
    \vspace{-5pt}
\end{figure*}

In Fig.~\ref{fig:appendix_visualize_query}, we visualize the results of 10 queries per sample, including detection boxes, detection scores, and referential scores. Additionally, we present the corresponding ground truth, predicted proposals, predicted referential targets, and referential masks.

\subsection{Detail Visualization}
\label{appendix:detail_visualization}
\begin{figure*}
    \centering
    \includegraphics[width=1.0\linewidth]{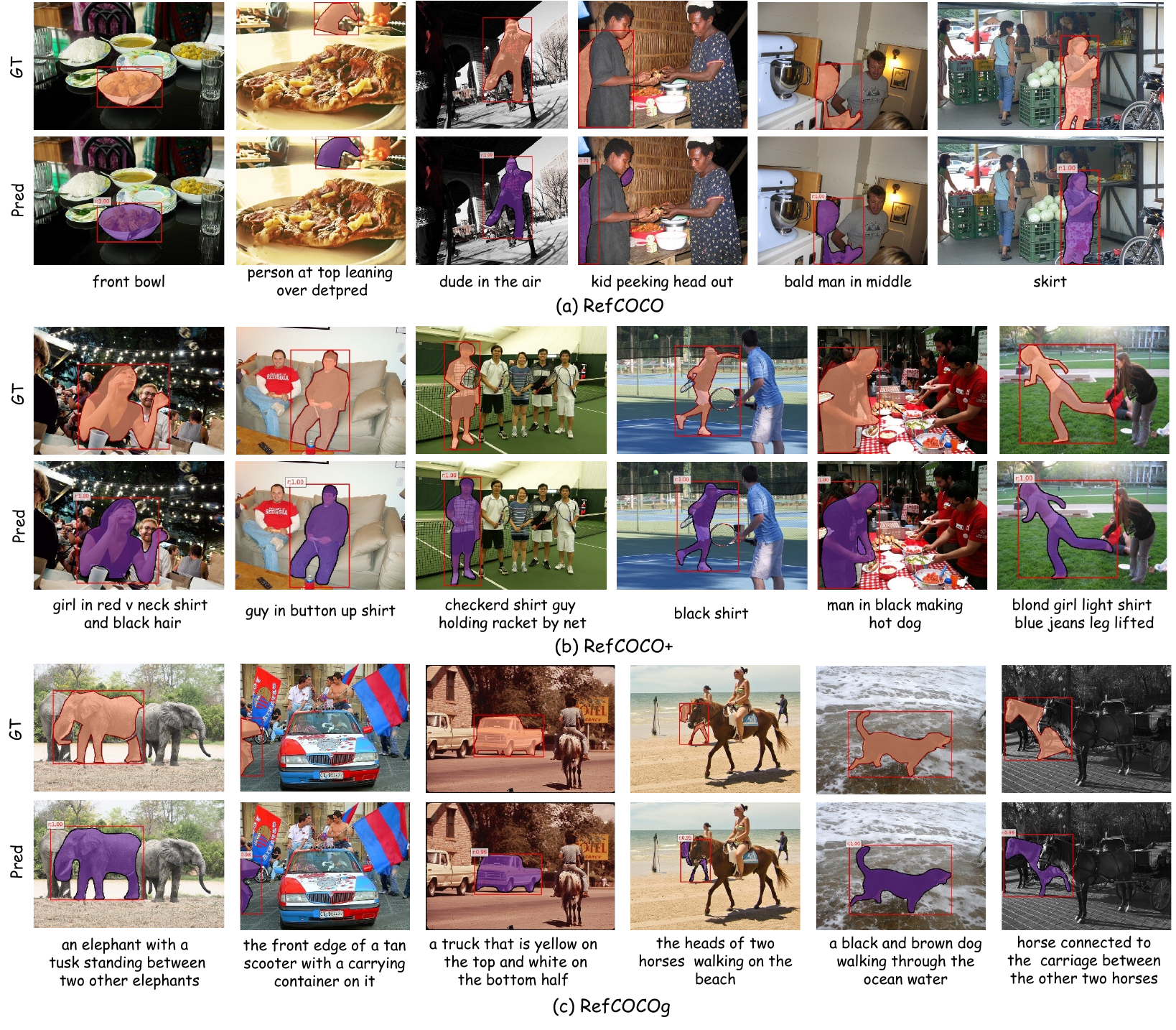}
    \vspace{-10pt}
    \caption{\textbf{Visualization of RefCOCO/+/g dataset.} }
    \label{fig:appendix_visualize_recoco}
   \vspace{-5pt}
\end{figure*}

\begin{figure*}
    \centering
    \includegraphics[width=1.0\linewidth]{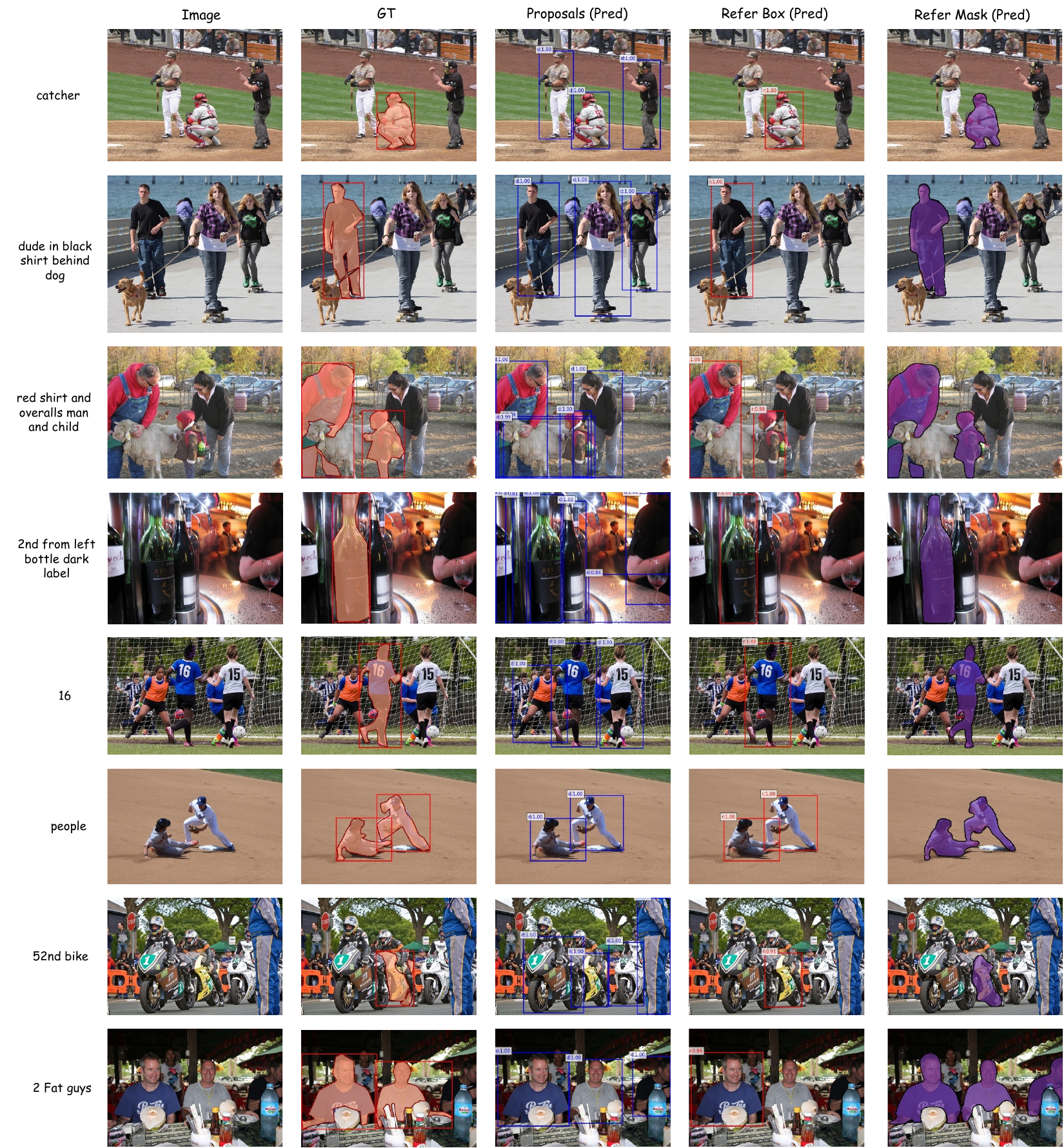}
    \vspace{-10pt}
    \caption{\textbf{Visualization of gRefCOCO dataset.} Proposals(Pred) refers to the predicted proposal boxes, Refer Box(Pred) refers to the predicted refer boxes, and Refer Mask(Pred) refers to the predicted refer masks.}
    \label{fig:appendix_visualize_grecoco}
   \vspace{-5pt}
\end{figure*}

\begin{figure*}
    \centering
    \includegraphics[width=1.0\linewidth]{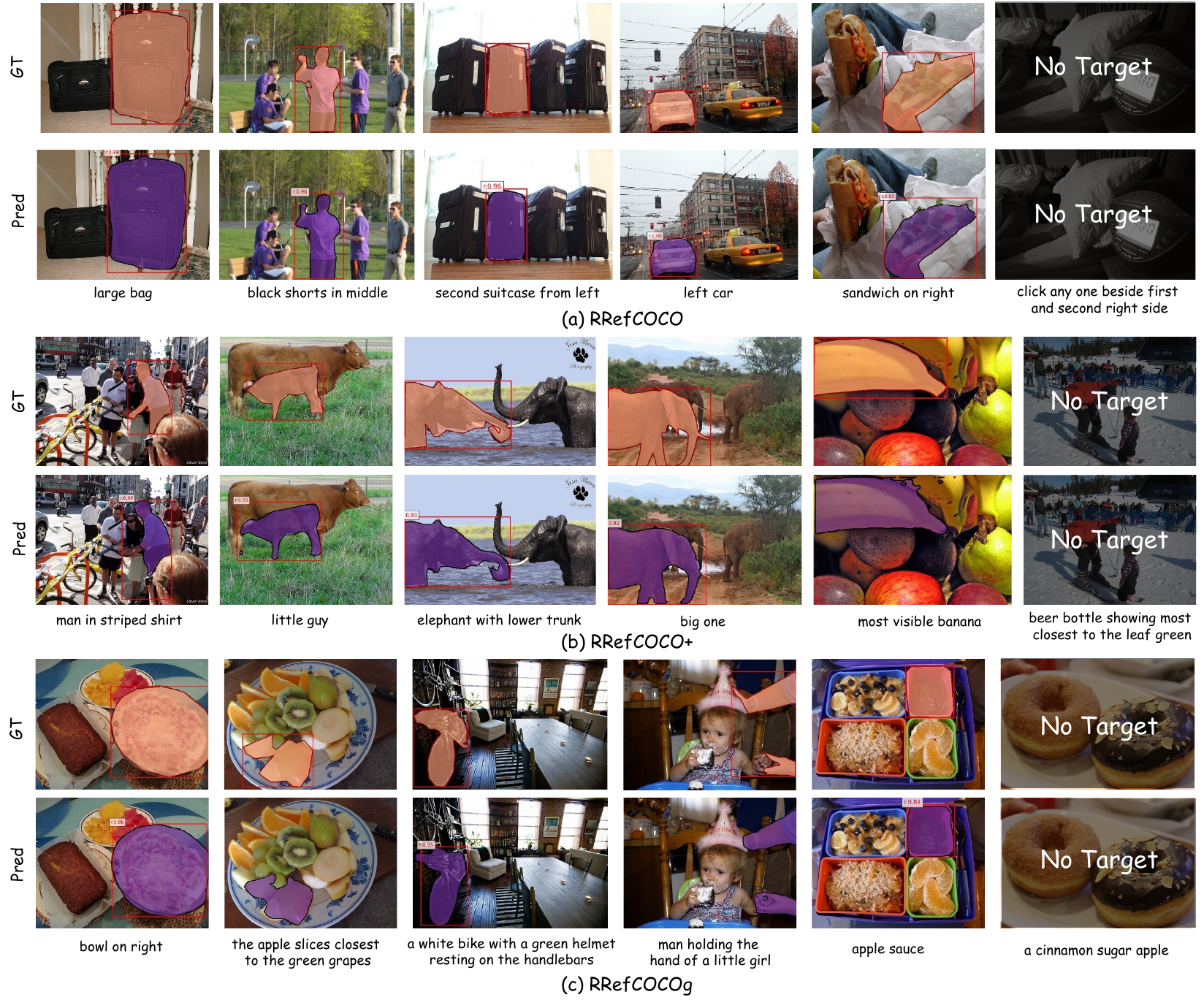}
    \vspace{-10pt}
    \caption{\textbf{Visualization of R-RefCOCO/+/g dataset.}}
    \label{fig:appendix_visualize_rrecoco}
   \vspace{-5pt}
\end{figure*}

\begin{figure*}
    \centering
    \includegraphics[width=1.0\linewidth]{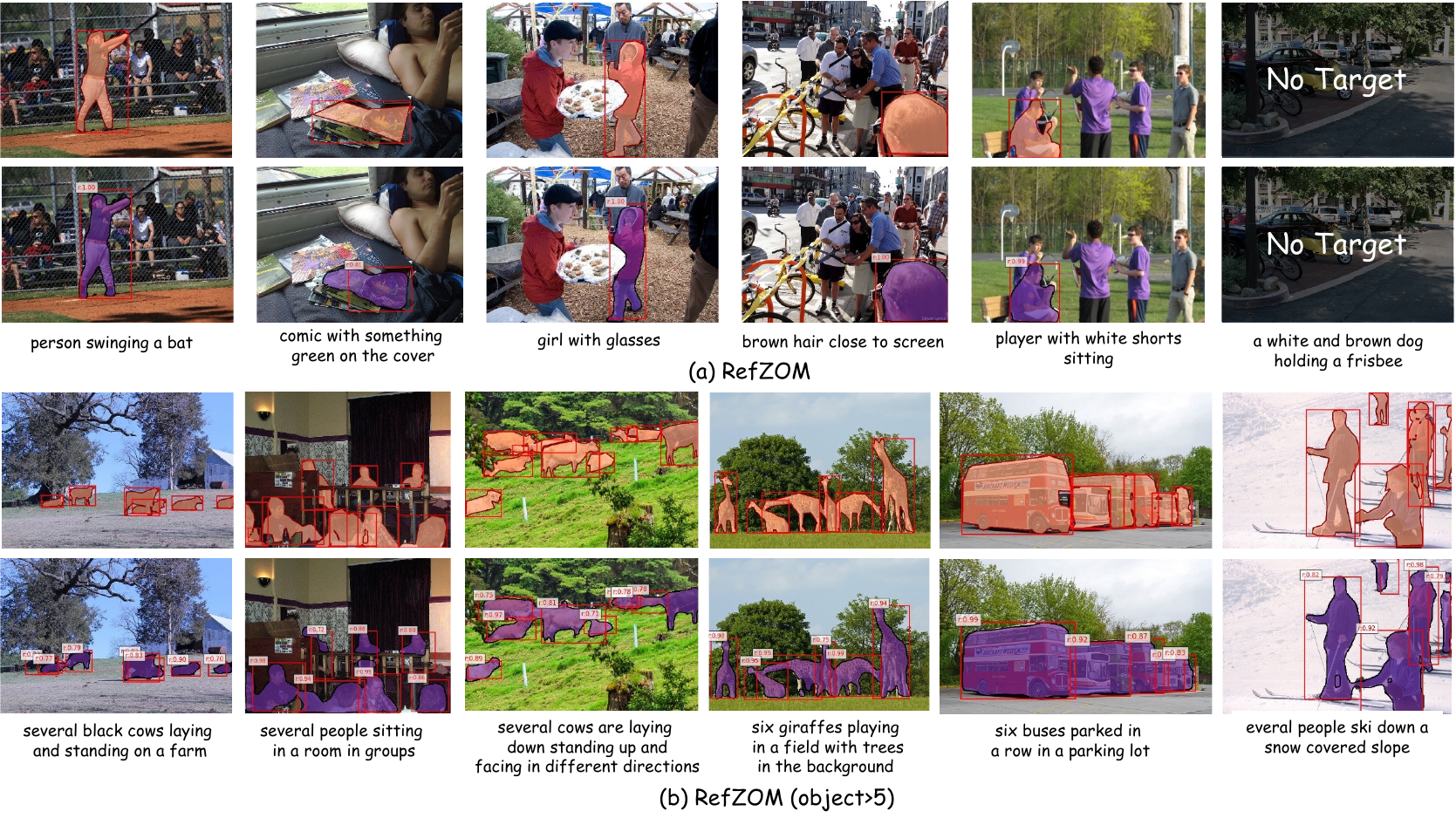}
    \vspace{-10pt}
    \caption{\textbf{Visualization of Ref-ZOM Dataset.} RefZOM (object$>$5) refers to samples in the Ref-ZOM dataset where the number of objects exceeds five.}
    \label{fig:appendix_visualize_refzom}
   \vspace{-5pt}
\end{figure*}

We visualize the detection and segmentation results on multiple datasets, including grefCOCO, RefCOCO/+/, R-RefCOCO/+/, and Ref-ZOM.
In Fig.~\ref{fig:appendix_visualize_recoco}, we present the detection and segmentation performance of PropVG on the standard RefCOCO dataset. Fig.~\ref{fig:appendix_visualize_grecoco} demonstrates the model's ability to extract foreground bounding boxes and resolve referential expressions. Fig.~\ref{fig:appendix_visualize_rrecoco} highlights PropVG’s performance in detection and segmentation under challenging scenarios, showcasing its robustness. Finally, Fig.~\ref{fig:appendix_visualize_refzom} illustrates the referential capability of PropVG on the Ref-ZOM dataset, including both standard and multi-referent cases. This also demonstrates the model’s enhanced detection robustness enabled by its foreground extraction capability.

\end{document}